\documentclass[letterpaper]{article} % DO NOT CHANGE THIS
\usepackage{aaai24}  % DO NOT CHANGE THIS
\usepackage{times}  % DO NOT CHANGE THIS
\usepackage{helvet}  % DO NOT CHANGE THIS
\usepackage{courier}  % DO NOT CHANGE THIS
\usepackage[hyphens]{url}  % DO NOT CHANGE THIS
\usepackage{graphicx} % DO NOT CHANGE THIS
\usepackage{natbib}  % DO NOT CHANGE THIS AND DO NOT ADD ANY OPTIONS TO IT
\usepackage{tikz}
\usepackage{amssymb}
\usepackage{amsmath}
\usepackage{caption} % DO NOT CHANGE THIS AND DO NOT ADD ANY OPTIONS TO IT
\usepackage{algorithm}
\usepackage{algorithmic}
\usepackage{newfloat}
\usepackage{listings}
\DeclareCaptionStyle{ruled}{labelfont=normalfont,labelsep=colon,strut=off} % DO NOT CHANGE THIS
\floatstyle{ruled}
\newfloat{listing}{tb}{lst}{}
\floatname{listing}{Listing}
\title{{4D-based Robot Navigation Using Relativistic Image Processing}}
\author{
    Simone Müller\textsuperscript{\rm 1},
    Dieter Kranzlmüller\textsuperscript{\rm 2}
}
\affiliations{
    \textsuperscript{\rm 1}Leibniz Supercomputing Centre\\
    \textsuperscript{\rm 2}Ludwig-Maximilians-Universität München\\
    simone.mueller@lrz.de, kranzlmueller@ifi.lmu.de
}

\begin{document}

\maketitle

\begin{abstract}
Machine perception is an important prerequisite for safe interaction and locomotion in dynamic environments. This requires not only the timely perception of surrounding geometries and distances but also the ability to react to changing situations through predefined, learned but also reusable skill endings of a robot so that physical damage or bodily harm can be avoided. In this context, 4D perception offers the possibility of predicting one's own position and changes in the environment over time. In this paper, we present a 4D-based approach to robot navigation using relativistic image processing. Relativistic image processing handles the temporal-related sensor information in a tensor model within a constructive 4D space. 4D-based navigation expands the causal understanding and the resulting interaction radius of a robot through the use of visual and sensory 4D information. 

\end{abstract}

% Hinter jeder Roboteraufgabe oder -interaktion steht eine Repräsentation, die (a) eine ausreichende Kontextualisierung ermöglicht, (b) alle vorhandenen vordefinierten, erlernten und/oder wiederverwendbaren Fähigkeiten an Bord des Roboters unterstützt, (c) zur Entwurfszeit überprüfbar ist und sich zur Laufzeit konsistent verhält und (d) für die Wiederverwendung auf einer Vielzahl unterschiedlicher Robotermorphologien getestet, ausgeführt und modifiziert werden kann. Den Endnutzern die Möglichkeit zu geben, ihre Absichten in verschiedenen Darstellungen auszudrücken, spielt seit langem eine zentrale Rolle bei der Entwicklung von Roboteranwendungen, d. h. bei der Konstruktion von Roboterdiensten, sozialen Interaktionen und/oder kollaborativen Aufgaben.
% Das Problem ist, dass es an Konsistenz und Einheitlichkeit bei der Auswahl und Verwendung dieser Darstellungen durch Robotikforscher mangelt. Daher stehen die Endnutzer (d. h. die Entwickler von Roboteranwendungen) aufgrund dieses Mangels an Kohärenz vor einer Vielzahl von Herausforderungen.

% Den kontext von informationen zu interpretieren und zu verstehen die aus einer zeitlichen und situativen abhängigkeit geschehen sind. 

\section{Introduction}
% Ausgangslage Allgemein

% KI-Planung für die Robotik, Roboterarchitekturen und Modelle für logisches Denken

% \cite{Aguinaldo.2023} - Bei der Planung von Roboteraufgaben kann die Verfolgung aller impliziten Effekte und Beziehungen im Weltzustand eine große Herausforderung darstellen, insbesondere bei der Arbeit in komplexen Umgebungen. Infolgedessen verlassen sich Planungssysteme oft auf Heuristiken und vereinfachende Annahmen, die zu suboptimalen oder sogar falschen Lösungen führen können (Wilkins und DesJardins 2001; Gil 1990). 

% \cite{s24082525} TD3 Algorithmus

Depth perception and corresponding spatial perception are essential for localization, navigation, obstacle avoidance, 3D mapping, and associated trajectory of motion in numerous areas of application, such as 3D modeling, mixed reality, autonomous vehicle systems, and robotics~\cite{SK.16,RSMM.13}. 

For example, a moving robot needs to observe its environment and be aware of objects and changes around it to make accurate predictions and decisions according to its tasks. In addition to depth perception, the robots' positioning and motion must be precise and predictable to react to changing situations and avoid physical harm. 

%Wie passt die Submission in den Workshop?
The path planning of robots and autonomous mobile robots (AMR) can be quite challenging in a dynamic environment where numerous conditions influencing visual perception, such as weather, lighting, as well as contrast, exist, and in which object and personal related distances in the form of motion depend on time. Requirements such as maintaining safety distances, collision avoidance due to moving objects, calculating the shortest possible times at minimal energy consumption must be considered~\cite{LOGANATHAN2023101343}. %In addition, individual sensors can exhibit inaccuracies, noise, consumption or integration errors, satellite signals cannot be used everywhere and several sensors must be correctly fused which can be a challenges~\cite{MK.22}. 

The positional inaccuracies of dynamic environments can also reinforce temporal inaccuracies or deviations of reference systems. In practice, GPS-based tracking systems (e.g., Garmin Oregon 700) exhibit position deviations of 3~m to 25~m \cite{gar22, MK.22}. Furthermore, the GPS signal does not extend sufficiently into tunnels or caves, making tracking difficult. This poses a problem because robotic systems are particularly suitable for remote or otherwise inaccessible areas due to their mobility, compactness, and robustness. The accuracy can be increased by using local signals to support the GPS signal. In the case of automated lawn mowers like Husquarna~\cite{husqvarna24}, the accuracy can be optimized to 1.5cm. Further, vehicles use the antennas of traffic to initialize their GPS signal. 

The use of global navigation of inertial navigation systems (GNSS/INS) enables a temporal position synchronization of these sensors \cite{8793604,10.1109/TITS.2012.2187641}. However, these positional deviations impair the quality of robot navigation. In addition, alternative visual-inertial navigation systems (VINS) offer limited advantages for multi-sensor alignment in terms of sensor deviations, measurement errors, range, and temporal effects \cite{car16,8793604,Kadambi20143DDC}.

To reduce the complexity of navigation in a dynamic environment various classical and heuristic techniques have been proposed. Thereby, path planning, which is considered a non-deterministic polynomial time (``NP'') problem~\cite{Erickson13}, becomes more complicated when the degrees of freedom of the system increase, e.g., when navigating in a 3-dimensional (3D) environment~\cite{LOGANATHAN2023101343}. Nevertheless, this form of navigation is necessary to adapt motions and temporal relations of information in a dynamic environment.  

%Was ist die Contribution?
Using 4D information in robot navigation offers significant advantages over 3D navigation. The position assignment in 4D navigation can be described using several rotations and translations that are related in time. This could favor navigation speed, as time delays in information, such as with GPS signals, could be predictively corrected over longer distances. 

Based on the challenges described above, we introduce a model for 4D-based navigation of robotic systems. Our evaluation reveals the feasibility and transferability of 4D sensor perception by using the principles of relativistic image processing. Our findings extend research on the benefits of sensor-based 4D Navigation and description of 4D motion. \\

This article contains the following {major contributions}:

\begin{itemize}
  \item{Spatio-temporal view of motion based on reference reconstruction}
  \item{4D-Positioning and design of 4D-based robot navigation}
  \item{Visualization of Schlingel coordinates and sensor maps}
\end{itemize}

% Was sind die Ergebnisse?
Our results demonstrate the effectiveness of 4D-based robot navigation by using the principles of relativistic image processing. Compared to the Euclidean locomotion of 6 degrees of freedom (DoFs), expressed by 3 translations and 3 rotations, the non-Euclidean representation of 4D space involves 10 DoFs. We use these relationships to describe the relativistic view of sensor-perceived as 4D Navigation in terms of 4 translations and 6 rotations.  

%Wer profitiert davon?

%The autonomous mobility and robotics sector comprises innovative technologies with a focus on increasing user comfort, safety and the ability to interact with their surroundings. By fused-using sensors, immediate surroundings can be acquired and spatially assessed. In contrast to conventional LiDAR and radar systems, cameras contain a lot of information and rich textures compressed into one image \cite{S.22}.

\section{Concept of 4D-based Robot Navigation}

Over the past years, numerous models have been developed to visualize higher-order time derivatives for predicting model-based control of robots and general description of multi-body systems by trajectory motion \cite{SK.16,D.22}. The point-based calculation of motion can be used to estimate positions, distances, and even areas or volumes. Information such as accelerations, orientations, or magnetic fields from different sensor types can be synchronized to determine positions, for example~\cite{MK.22}.

\vspace*{-0.3cm}

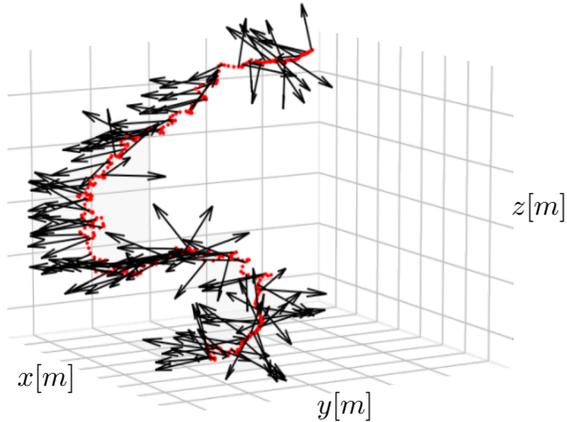
\begin{figure}[ht!]
    \centering

\tikzset{every picture/.style={line width=0.75pt}} %set default line width to 0.75pt        

\begin{tikzpicture}[x=0.75pt,y=0.75pt,yscale=-1,xscale=1]
%uncomment if require: \path (0,10069); %set diagram left start at 0, and has height of 10069

%Image [id:dp5240918514671391] 
\draw (645.18,1795.6) node  {\includegraphics[width=208.77pt,height=160.71pt]{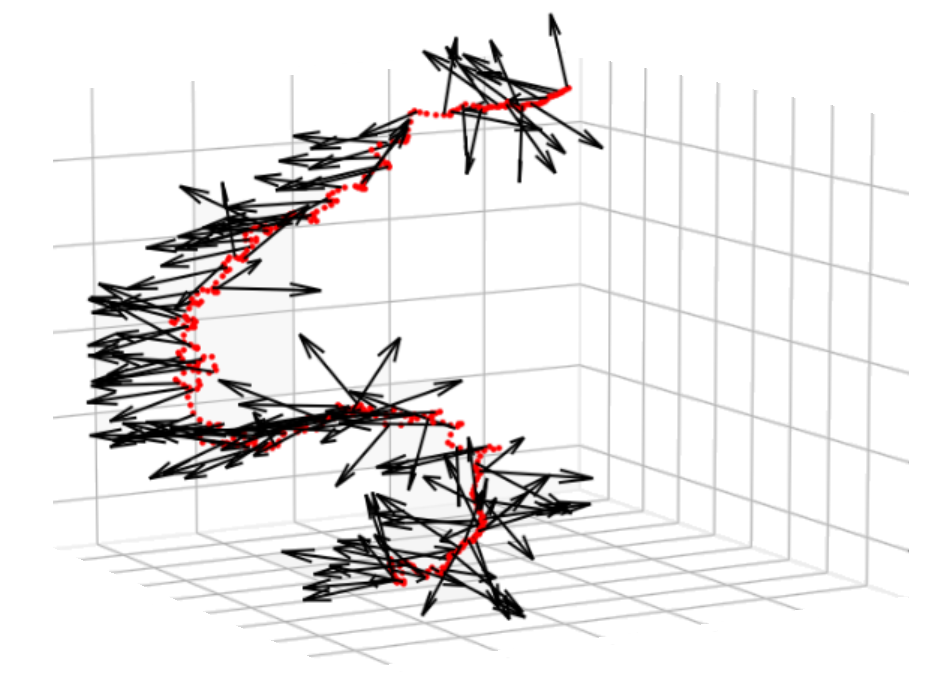}};

% Text Node
\draw (774,1787) node [anchor=north west][inner sep=0.75pt]  [font=\large] [align=left] {$z [m]$};
% Text Node
\draw (676,1887) node [anchor=north west][inner sep=0.75pt]  [font=\large] [align=left] {$y [m]$};
% Text Node
\draw (525,1873) node [anchor=north west][inner sep=0.75pt]  [font=\large] [align=left] {$x [m]$};
\end{tikzpicture}

    \caption{ The trajectory, adapted and reproduced from~\cite{MK.22}, demonstrates the sensor-acquired climbing of stairs in an Euclidean diagram. The black arrows represent the sensor orientation, the red dots represent the calculated position. }
    \label{fig:TrajectoryIllustration}
\end{figure}

Conventional trajectories, as shown in the example of Figure~\ref{fig:TrajectoryIllustration}, can represent changes in position (red dots) and vectorial orientations (black arrows). Basically, movement in Euclidean space is described by 6 DoF. The temporal sequence and representation of this information can only be logically assumed as there is no temporal axis.

%Prior research indicates that minor deviations of aligned point clouds can be minimised by accurate positional information. Data fusion of inertial sensors in form of smart sensor architectures reduce the influence of a failure prone single sensor \cite{car16}.

% \section{Concept of 4D-based Robot Navigation}

In our concept, we define a {space of 4 dimensions} that includes time as part of metric space $(\forall$ $x,y,z,\zeta \in$ $\chi^{\mu}$)\label{eq:MetricSpaceV1}. In this space, $\zeta$ describes the spatial coordinate of time. The speed of light $c [m/s]$ and time $\tau [s]$ are set as location $\zeta [m]$ further spatial axis.
The local axis $\zeta [m \cdot s^{-1} \cdot s = m ]$ involves a speed of light of $c=1m/s$ through the unit reduction~\cite{Farook2022}. Setting the speed of light to the natural unit ($c=1$) is only valid in 4D space $\mathbb{R}^{4}$. We consider $c$ as a natural conversion factor for the discrete space and time components. This allows the verification of temporal changes as a meter unit. Furthermore, it simplifies the use of relativistic equations and enables the transferability to different sensor types, since space and time are measured in the same unit. 

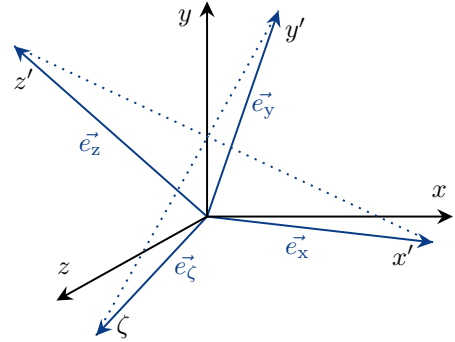
\begin{figure}[ht!]
    \centering

    \tikzset{every picture/.style={line width=0.75pt}} %set default line width to 0.75pt        

\tikzset{every picture/.style={line width=0.75pt}} %set default line width to 0.75pt        

\begin{tikzpicture}[x=0.75pt,y=0.75pt,yscale=-1,xscale=1]
%uncomment if require: \path (0,10069); %set diagram left start at 0, and has height of 10069

%Straight Lines [id:da11187301217545409] 
\draw [color={rgb, 255:red, 9; green, 60; blue, 134 }  ,draw opacity=1 ][line width=0.75]    (442.15,7911.96) -- (477.18,7810.87) ;
\draw [shift={(478.17,7808.03)}, rotate = 109.12] [fill={rgb, 255:red, 9; green, 60; blue, 134 }  ,fill opacity=1 ][line width=0.08]  [draw opacity=0] (8.04,-3.86) -- (0,0) -- (8.04,3.86) -- (5.34,0) -- cycle    ;
%Straight Lines [id:da44164080493817903] 
\draw [color={rgb, 255:red, 9; green, 60; blue, 134 }  ,draw opacity=1 ][line width=0.75]    (442.15,7911.96) -- (553.19,7924.69) ;
\draw [shift={(556.17,7925.03)}, rotate = 186.54] [fill={rgb, 255:red, 9; green, 60; blue, 134 }  ,fill opacity=1 ][line width=0.08]  [draw opacity=0] (8.04,-3.86) -- (0,0) -- (8.04,3.86) -- (5.34,0) -- cycle    ;

%Straight Lines [id:da20927504834219857] 
\draw [color={rgb, 255:red, 9; green, 60; blue, 134 }  ,draw opacity=1 ][line width=0.75]    (442.15,7911.96) -- (388.21,7969.84) ;
\draw [shift={(386.17,7972.03)}, rotate = 312.98] [fill={rgb, 255:red, 9; green, 60; blue, 134 }  ,fill opacity=1 ][line width=0.08]  [draw opacity=0] (8.04,-3.86) -- (0,0) -- (8.04,3.86) -- (5.34,0) -- cycle    ;

%Straight Lines [id:da5950932338682149] 
\draw [color={rgb, 255:red, 9; green, 60; blue, 134 }  ,draw opacity=1 ][line width=0.75]    (442.15,7911.96) -- (347.05,7827.92) ;
\draw [shift={(344.8,7825.94)}, rotate = 41.47] [fill={rgb, 255:red, 9; green, 60; blue, 134 }  ,fill opacity=1 ][line width=0.08]  [draw opacity=0] (8.04,-3.86) -- (0,0) -- (8.04,3.86) -- (5.34,0) -- cycle    ;
%Straight Lines [id:da6953537376939154] 
\draw [color={rgb, 255:red, 0; green, 0; blue, 0 }  ,draw opacity=1 ][line width=0.75]    (442.15,7911.96) -- (563.17,7911.96) ;
\draw [shift={(566.17,7911.96)}, rotate = 180] [fill={rgb, 255:red, 0; green, 0; blue, 0 }  ,fill opacity=1 ][line width=0.08]  [draw opacity=0] (8.04,-3.86) -- (0,0) -- (8.04,3.86) -- (5.34,0) -- cycle    ;
%Straight Lines [id:da8244467954290635] 
\draw [color={rgb, 255:red, 0; green, 0; blue, 0 }  ,draw opacity=1 ][line width=0.75]    (442.15,7911.96) -- (442.15,7806.36) ;
\draw [shift={(442.15,7803.36)}, rotate = 90] [fill={rgb, 255:red, 0; green, 0; blue, 0 }  ,fill opacity=1 ][line width=0.08]  [draw opacity=0] (8.04,-3.86) -- (0,0) -- (8.04,3.86) -- (5.34,0) -- cycle    ;
%Straight Lines [id:da07711461713008094] 
\draw [color={rgb, 255:red, 0; green, 0; blue, 0 }  ,draw opacity=1 ][line width=0.75]    (442.15,7911.96) -- (368.79,7952.9) ;
\draw [shift={(366.17,7954.36)}, rotate = 330.83] [fill={rgb, 255:red, 0; green, 0; blue, 0 }  ,fill opacity=1 ][line width=0.08]  [draw opacity=0] (8.04,-3.86) -- (0,0) -- (8.04,3.86) -- (5.34,0) -- cycle    ;
%Straight Lines [id:da8304009826500118] 
\draw [color={rgb, 255:red, 9; green, 60; blue, 134 }  ,draw opacity=1 ] [dash pattern={on 0.84pt off 2.51pt}]  (344.8,7825.94) -- (556.17,7925.03) ;
%Straight Lines [id:da7837916996516117] 
\draw [color={rgb, 255:red, 9; green, 60; blue, 134 }  ,draw opacity=1 ] [dash pattern={on 0.84pt off 2.51pt}]  (386.17,7972.03) -- (478.17,7808.03) ;

% Text Node
\draw (395,7960) node [anchor=north west][inner sep=0.75pt]   [align=left] {$\zeta$};
% Text Node
\draw (534,7925) node [anchor=north west][inner sep=0.75pt]   [align=left] {$x'$};
% Text Node
\draw (480,7810) node [anchor=north west][inner sep=0.75pt]   [align=left] {$y'$};
% Text Node
\draw (343,7836) node [anchor=north west][inner sep=0.75pt]   [align=left] {$z'$};
% Text Node
\draw (426,7806) node [anchor=north west][inner sep=0.75pt]   [align=left] {$y$};
% Text Node
\draw (365,7934) node [anchor=north west][inner sep=0.75pt]   [align=left] {$z$};
% Text Node
\draw (424,7930) node [anchor=north west][inner sep=0.75pt]   [color={rgb, 255:red, 9; green, 60; blue, 134 }  ,opacity=1 ][align=left] {$\vec{e_{\zeta}}$};
% Text Node
\draw (480,7920) node [anchor=north west][inner sep=0.75pt]  [color={rgb, 255:red, 9; green, 60; blue, 134 }  ,opacity=1 ] [align=left] {$\vec{e_\mathrm{x}}$};
% Text Node
\draw (376,7868.14) node [anchor=north west][inner sep=0.75pt] [color={rgb, 255:red, 9; green, 60; blue, 134 }  ,opacity=1 ]  [align=left] {$\vec{e_\mathrm{z}}$};
% Text Node
\draw (463,7848) node [anchor=north west][inner sep=0.75pt]   [color={rgb, 255:red, 9; green, 60; blue, 134 }  ,opacity=1 ][align=left] {$\vec{e_\mathrm{y}}$};
% Text Node
\draw (554,7895) node [anchor=north west][inner sep=0.75pt]   [align=left] {$x$};

\end{tikzpicture}
    
    \caption{This figure illustrates the tensor-based reference diagram. The novel 4D diagram converts Euclidean coordinates into spatio-temporal coordinates by composing tensor bases.}
    \label{fig:SchlingelTensorBases}
\end{figure}

The reference bases of Figure~\ref{fig:SchlingelTensorBases} are an intersection between the linearly independent sensor data and a generating system of relativistic information. The Schlingel notation corresponds to the notation of 4D space and therefore constitutes 4 spatial axes: $\zeta,x,y,z$. The {metric space} in our Schlingel diagram shall be defined as $d: M \times M \rightharpoonup \mathbb{R}$ with the conventional 2-order tensor $\mu: V \times V \rightharpoonup \mathbb{R}$, $\mathbb{R}^{1,4}: (\mathbb{R},\chi^{\mu})$.

%Wir stellen das {Schlingel-Diagramm} als einen neuen Ansatz vor um zeit- und raumabhängige Bewegungen durch 10 DoF darzustellen. 

\subsection{Fundamentals of Schlingel Diagram}
In this section, we define a coordinate system that contains spatial and temporal information in order to assign fixed coordinates for motion and navigation. We introduce the {Schlingel diagram} as a novel approach to visualize space- and time-dependent motion.  

\vspace*{0.3cm}

\textbf{Definition:}~\textit{The Schlingel diagram defines a novel approach of a coordinate system in which space- and time-dependent information are expressed by 10 different degrees of freedom (DoF), consisting of 4 translations and 6 rotations. It provides a new basis for relating different types of information to each other, considering temporal aspects. }

\vspace*{0.3cm}
% It can be used to visualize dependencies ...

Previous approaches of visualizing higher dimensional information, such as Minkowski diagram~\cite{BMW.22} or mathematical subspaces, often prove to be complex and cluttered. In this context, the Schlingel diagram offers a new and clearer approach where the space and time-related aspects of information are summarized as a spatio-temporal coordinate (Schlingel coordinate). The limits of the Schlingel diagram are still undergoing technical validation for further applications. In this article, we derive the 4D position using the Schlingel diagram. 

The contained quantity of the environment $\epsilon$ in the Schlingel related vector space $V_{\epsilon}(x_\mathrm{0})$$:= \{ x \in \chi ~ | ~ d(x,x_\mathrm{0}) < \epsilon \}$ requires a metric, which we define as a metric space ($\chi,d$) with $x_\mathrm{0} \in \chi $ and $\epsilon > 0$. The motion in space is defined as {four-vector} $\chi^{\mu}$ and refers to  Einstein's summation convention $(+---)$ \cite{E.16}:
\begin{equation}\label{eq:EinsteinSummenkonvention} 
    d s^{2} = d \zeta^{2} - d x^{2} - d y^{2} - d z^{2} \hspace*{0.0cm} \in \mathbb{R}^{4} \hspace*{0.3cm} with \hspace*{0.3cm} \zeta^{2} = c^{2} d t^{2}
\end{equation}
Where $\chi^{\mu} = (\chi^{0}+\chi^{1}+\chi^{2}+\chi^{3})$ contains the following form:
  \begin{equation}\label{eq:ViererVektor} 
       \chi^{\mu} = ~A^{\zeta} \vec{e_\mathrm{\zeta}} + A^{x} \vec{e_\mathrm{x}} + A^{y} \vec{e_\mathrm{y}} + A^{z} \vec{e_\mathrm{z}} \hspace*{0.2cm} \{ \hspace*{0.1cm}  \chi^{\mu} = [\zeta, x,y,z] \in \mathbb{R} 
    \end{equation}

The shown Schlingel diagram in Fig.~\ref{fig:DynamicStateSchlingel} introduces the existing spatio-temporal rotations, translations, and resulting four-vector $\chi^{\mu}$.  

 \begin{figure}[ht!]
     \centering

\tikzset{every picture/.style={line width=0.75pt}} %set default line width to 0.75pt        

\begin{tikzpicture}[x=0.75pt,y=0.75pt,yscale=-1,xscale=1]
%uncomment if require: \path (0,23566); %set diagram left start at 0, and has height of 23566

%Shape: Right Triangle [id:dp4061202464450475] 
\draw  [draw opacity=0][fill={rgb, 255:red, 255; green, 255; blue, 255 }   ,fill opacity=0.02 ] (371.69,11969.04) -- (364.01,12121.17) -- (266.75,12081.34) -- cycle ;
%Shape: Right Triangle [id:dp3806853789260116] 
\draw  [draw opacity=0][fill={rgb, 255:red, 255; green, 255; blue, 255 }   ,fill opacity=0.02 ] (376.48,11962.17) -- (197.78,11955.56) -- (265.15,12079.74) -- cycle ;
%Shape: Right Triangle [id:dp5378895565907651] 
\draw  [draw opacity=0][fill={rgb, 255:red, 74; green, 74; blue, 74 }  ,fill opacity=0.09 ] (131.46,12083.2) -- (375.34,11962.94) -- (266.8,12080.2) -- cycle ;
%Straight Lines [id:da15575264185948656] 
\draw [color={rgb, 255:red, 0; green, 0; blue, 0 }  ,draw opacity=1 ][line width=0.75]    (265.15,12081.74) -- (198.23,11958.36) ;
\draw [shift={(196.8,11955.72)}, rotate = 61.53] [fill={rgb, 255:red, 0; green, 0; blue, 0 }  ,fill opacity=1 ][line width=0.08]  [draw opacity=0] (8.04,-3.86) -- (0,0) -- (8.04,3.86) -- (5.34,0) -- cycle    ;
%Straight Lines [id:da5542981493610848] 
\draw [color={rgb, 255:red, 0; green, 0; blue, 0 }  ,draw opacity=1 ][line width=0.75]    (265.15,12081.74) -- (373.76,11964.92) ;
\draw [shift={(375.8,11962.72)}, rotate = 132.91] [fill={rgb, 255:red, 0; green, 0; blue, 0 }  ,fill opacity=1 ][line width=0.08]  [draw opacity=0] (8.04,-3.86) -- (0,0) -- (8.04,3.86) -- (5.34,0) -- cycle    ;
%Straight Lines [id:da0662786638402666] 
\draw [color={rgb, 255:red, 0; green, 0; blue, 0 }  ,draw opacity=1 ][line width=0.75]    (265.15,12081.74) -- (360.03,12121.56) ;
\draw [shift={(362.8,12122.72)}, rotate = 202.76] [fill={rgb, 255:red, 0; green, 0; blue, 0 }  ,fill opacity=1 ][line width=0.08]  [draw opacity=0] (8.04,-3.86) -- (0,0) -- (8.04,3.86) -- (5.34,0) -- cycle    ;
%Straight Lines [id:da7044998599659238] 
\draw [color={rgb, 255:red, 0; green, 0; blue, 0 }  ,draw opacity=1 ][line width=0.75]    (265.15,12081.74) -- (135.8,12083.67) ;
\draw [shift={(132.8,12083.72)}, rotate = 359.14] [fill={rgb, 255:red, 0; green, 0; blue, 0 }  ,fill opacity=1 ][line width=0.08]  [draw opacity=0] (8.04,-3.86) -- (0,0) -- (8.04,3.86) -- (5.34,0) -- cycle    ;
%Straight Lines [id:da09991606940811071] 
\draw [color={rgb, 255:red, 0; green, 0; blue, 0 }  ,draw opacity=1 ][line width=0.75]  [dash pattern={on 0.84pt off 2.51pt}]  (362.8,12122.72) -- (132.8,12083.72) ;
%Straight Lines [id:da8074127898959136] 
\draw [color={rgb, 255:red, 0; green, 0; blue, 0 }  ,draw opacity=1 ][line width=0.75]  [dash pattern={on 0.84pt off 2.51pt}]  (362.8,12122.72) -- (375.8,11962.72) ;
%Straight Lines [id:da4213562891689646] 
\draw [color={rgb, 255:red, 0; green, 0; blue, 0 }  ,draw opacity=1 ][line width=0.75]  [dash pattern={on 0.84pt off 2.51pt}]  (375.8,11962.72) -- (196.8,11955.72) ;
%Straight Lines [id:da4889857583964572] 
\draw [color={rgb, 255:red, 0; green, 0; blue, 0 }  ,draw opacity=1 ][line width=0.75]  [dash pattern={on 0.84pt off 2.51pt}]  (132.8,12083.72) -- (196.8,11955.72) ;

%Shape: Circle [id:dp7801284513423665] 
\draw  [draw opacity=0][fill={rgb, 255:red, 171; green, 40; blue, 40 }  ,fill opacity=1 ] (263.55,12081.34) .. controls (263.55,12080.46) and (264.26,12079.74) .. (265.15,12079.74) .. controls (266.03,12079.74) and (266.75,12080.46) .. (266.75,12081.34) .. controls (266.75,12082.22) and (266.03,12082.94) .. (265.15,12082.94) .. controls (264.26,12082.94) and (263.55,12082.22) .. (263.55,12081.34) -- cycle ;

%Shape: Circle [id:dp1981085564229832] 
\draw  [draw opacity=0][fill={rgb, 255:red, 2; green, 0; blue, 214 }  ,fill opacity=1 ] (284.18,11977.09) .. controls (284.18,11975.98) and (285.08,11975.09) .. (286.18,11975.09) .. controls (287.28,11975.09) and (288.18,11975.98) .. (288.18,11977.09) .. controls (288.18,11978.19) and (287.28,11979.09) .. (286.18,11979.09) .. controls (285.08,11979.09) and (284.18,11978.19) .. (284.18,11977.09) -- cycle ;
%Straight Lines [id:da9775926986692474] 
\draw [color={rgb, 255:red, 254; green, 0; blue, 0 }  ,draw opacity=1 ]   (180.63,12013.71) -- (233.87,11958.36) ;
\draw [shift={(235.95,11956.2)}, rotate = 133.89] [fill={rgb, 255:red, 254; green, 0; blue, 0 }  ,fill opacity=1 ][line width=0.08]  [draw opacity=0] (7.14,-3.43) -- (0,0) -- (7.14,3.43) -- (4.74,0) -- cycle    ;
%Straight Lines [id:da028863169062404603] 
\draw [color={rgb, 255:red, 254; green, 0; blue, 0 }  ,draw opacity=1 ]   (235.95,11956.2) -- (280.44,11975.87) ;
\draw [shift={(283.18,11977.09)}, rotate = 203.86] [fill={rgb, 255:red, 254; green, 0; blue, 0 }  ,fill opacity=1 ][line width=0.08]  [draw opacity=0] (7.14,-3.43) -- (0,0) -- (7.14,3.43) -- (4.74,0) -- cycle    ;
%Straight Lines [id:da6892625453545824] 
\draw [color={rgb, 255:red, 254; green, 0; blue, 0 }  ,draw opacity=1 ]   (265.15,12081.74) -- (221,12081.74) ;
\draw [shift={(218,12081.74)}, rotate = 360] [fill={rgb, 255:red, 254; green, 0; blue, 0 }  ,fill opacity=1 ][line width=0.08]  [draw opacity=0] (7.14,-3.43) -- (0,0) -- (7.14,3.43) -- (4.74,0) -- cycle    ;
%Straight Lines [id:da8548234020236014] 
\draw [color={rgb, 255:red, 254; green, 0; blue, 0 }  ,draw opacity=1 ]   (218.97,12082.73) -- (182.08,12016.33) ;
\draw [shift={(180.63,12013.71)}, rotate = 60.94] [fill={rgb, 255:red, 254; green, 0; blue, 0 }  ,fill opacity=1 ][line width=0.08]  [draw opacity=0] (7.14,-3.43) -- (0,0) -- (7.14,3.43) -- (4.74,0) -- cycle    ;
%Straight Lines [id:da24872867252757325] 
\draw [color={rgb, 255:red, 0; green, 0; blue, 0 }  ,draw opacity=1 ][line width=1.5]  [dash pattern={on 5.63pt off 4.5pt}]  (265.15,12079.74) -- (285.36,11983) ;
\draw [shift={(286.18,11979.09)}, rotate = 101.8] [fill={rgb, 255:red, 0; green, 0; blue, 0 }  ,fill opacity=1 ][line width=0.08]  [draw opacity=0] (8.75,-4.2) -- (0,0) -- (8.75,4.2) -- (5.81,0) -- cycle    ;
%Straight Lines [id:da9461396689487034] 
\draw [color={rgb, 255:red, 0; green, 0; blue, 0 }  ,draw opacity=1 ][line width=0.75]  [dash pattern={on 4.5pt off 4.5pt}]  (287.15,11976.74) -- (336,11926.05) ;
%Straight Lines [id:da7123279171899712] 
\draw [color={rgb, 255:red, 0; green, 0; blue, 0 }  ,draw opacity=1 ][line width=0.75]  [dash pattern={on 4.5pt off 4.5pt}]  (285.15,11976.74) -- (237.5,11976.74) ;
%Straight Lines [id:da763559792219431] 
\draw [color={rgb, 255:red, 0; green, 0; blue, 0 }  ,draw opacity=1 ][line width=0.75]  [dash pattern={on 4.5pt off 4.5pt}]  (285.15,11976.74) -- (252,11914.05) ;
%Straight Lines [id:da5821065468881359] 
\draw [color={rgb, 255:red, 0; green, 0; blue, 0 }  ,draw opacity=1 ][line width=0.75]  [dash pattern={on 4.5pt off 4.5pt}]  (285.15,11976.74) -- (334,11998.05) ;
%Curve Lines [id:da44486947610647976] 
\draw    (257,11980.72) .. controls (268.4,12003.52) and (292.44,12002.85) .. (307.67,11992.44) ;
\draw [shift={(310,11990.72)}, rotate = 141.34] [fill={rgb, 255:red, 0; green, 0; blue, 0 }  ][line width=0.08]  [draw opacity=0] (7.14,-3.43) -- (0,0) -- (7.14,3.43) -- (4.74,0) -- cycle    ;
%Curve Lines [id:da28385029084893976] 
\draw    (316,11984.72) .. controls (319.52,11975.92) and (316.07,11965.57) .. (313.15,11958.44) ;
\draw [shift={(312,11955.72)}, rotate = 66.8] [fill={rgb, 255:red, 0; green, 0; blue, 0 }  ][line width=0.08]  [draw opacity=0] (7.14,-3.43) -- (0,0) -- (7.14,3.43) -- (4.74,0) -- cycle    ;
%Curve Lines [id:da6240396864933413] 
\draw    (307,11952.72) .. controls (299.76,11947.29) and (288.42,11945.95) .. (279.67,11949.46) ;
\draw [shift={(277,11950.72)}, rotate = 330.95] [fill={rgb, 255:red, 0; green, 0; blue, 0 }  ][line width=0.08]  [draw opacity=0] (7.14,-3.43) -- (0,0) -- (7.14,3.43) -- (4.74,0) -- cycle    ;
%Curve Lines [id:da8603086138822669] 
\draw    (269,11949.72) .. controls (265.36,11952.45) and (258.41,11959.32) .. (256.43,11968.82) ;
\draw [shift={(256,11971.72)}, rotate = 275.19] [fill={rgb, 255:red, 0; green, 0; blue, 0 }  ][line width=0.08]  [draw opacity=0] (7.14,-3.43) -- (0,0) -- (7.14,3.43) -- (4.74,0) -- cycle    ;
%Curve Lines [id:da5404825566295779] 
\draw    (319,11938.72) .. controls (310.82,11929.36) and (247.93,11919.03) .. (241.26,11969.38) ;
\draw [shift={(241,11971.72)}, rotate = 275.39] [fill={rgb, 255:red, 0; green, 0; blue, 0 }  ][line width=0.08]  [draw opacity=0] (7.14,-3.43) -- (0,0) -- (7.14,3.43) -- (4.74,0) -- cycle    ;
%Curve Lines [id:da9524805775814797] 
\draw    (328.3,11994.22) .. controls (330.91,11931.01) and (285.37,11908.31) .. (262.41,11919.38) ;
\draw [shift={(260,11920.72)}, rotate = 327.53] [fill={rgb, 255:red, 0; green, 0; blue, 0 }  ][line width=0.08]  [draw opacity=0] (7.14,-3.43) -- (0,0) -- (7.14,3.43) -- (4.74,0) -- cycle    ;
%Shape: Right Triangle [id:dp7054736879194325] 
\draw  [draw opacity=0][fill={rgb, 255:red, 255; green, 255; blue, 255 }  ,fill opacity=0.02 ] (266.75,12082.75) -- (131.2,12083.91) -- (197.48,11956.63) -- cycle ;
%Shape: Right Triangle [id:dp4248863159075691] 
\draw  [draw opacity=0][fill={rgb, 255:red, 255; green, 255; blue, 255 }  ,fill opacity=0.02 ] (361.2,12122.91) -- (131.2,12083.91) -- (268.99,12082.68) -- cycle ;
%Shape: Right Triangle [id:dp9265157533532415] 
\draw  [draw opacity=0][fill={rgb, 255:red, 74; green, 74; blue, 74 }  ,fill opacity=0.1 ] (197.51,11957.53) -- (356.79,12119.29) -- (267.15,12082.74) -- cycle ;

% Text Node
\draw (117,12076.72) node [anchor=north west][inner sep=0.75pt]   [align=left] {$\zeta$};
% Text Node
\draw (362,12120) node [anchor=north west][inner sep=0.75pt]   [align=left] {$x$};
% Text Node
\draw (375,11945.72) node [anchor=north west][inner sep=0.75pt]   [align=left] {$y$};
% Text Node
\draw (186,11942) node [anchor=north west][inner sep=0.75pt]   [align=left] {$z$};
% Text Node
\draw (340,11995) node [anchor=north west][inner sep=0.75pt]   [align=left] {$x'$};
% Text Node
\draw (334,11906.72) node [anchor=north west][inner sep=0.75pt]   [align=left] {$y'$};
% Text Node
\draw (241,11900) node [anchor=north west][inner sep=0.75pt]   [align=left] {$z'$};
% Text Node
\draw (226,11966.72) node [anchor=north west][inner sep=0.75pt]   [align=left] {$\zeta$};
% Text Node
%\draw (191,11919.05) node [anchor=north west][inner sep=0.75pt]   [align=left] {$p$};

% Text Node
\draw (279,12021.72) node [anchor=north west][inner sep=0.75pt]  [color={rgb, 255:red, 0; green, 0; blue, 0 }  ,opacity=1 ] [align=left] {\textbf{$\chi^{\mu}$}};

% Text Node
\draw (192,12005.72) node [anchor=north west][inner sep=0.75pt]  [color={rgb, 255:red, 0; green, 0; blue, 0 }  ,opacity=1 ] [align=left] {\textcolor[rgb]{0.89,0.05,0.05}{$\mathcal{T}_{4}^{1,4}$}};
% Text Node
\draw (297,11902) node [anchor=north west][inner sep=0.75pt]  [color={rgb, 255:red, 0; green, 0; blue, 0 }  ,opacity=1 ] [align=left] {$\mathcal{R}_{6}^{1,4}$};

\end{tikzpicture}
     \caption{ The figure illustrates 10-DoF Motion in the Model of Schlingel Diagram. In the dynamic state, motion occurs within all 6 planes. The motion itself is demonstrated by 4 translations and 6 rotations. }
     \label{fig:DynamicStateSchlingel}
 \end{figure}

The resulting variable $\chi^{\mu}$ contains 10 degrees of freedom consisting of 4 translations [$T_{\zeta},T_{x},T_{y},T_{z}$] and 6 rotations [$\mathcal{R}_{\zeta, x},\mathcal{R}_{\zeta,y},\mathcal{R}_{\zeta,z},\mathcal{R}_{x,y},\mathcal{R}_{x,z},\mathcal{R}_{y,z}$]. We express $\chi^{\mu}$ as follow:  

\vspace*{-0.4cm}

 \begin{equation}\label{eq:Spatio-Temoral-Motion}
   \chi^{\mu} = (\mathcal{R}_\mathrm{6}|\mathcal{T}_\mathrm{4}) e_{\tau} \in \mathbb{R}^{1,4}
\end{equation}

\vspace*{0.2cm}
 
Applying the four-vector allows the relativistic view of successive movements. To convert one reference system into another or to compare reference systems with each other, it is necessary to apply the \textbf{Lorentz transformation} $\Gamma$. The Lorentz transformation relates temporal and spatial coordinates of one reference system to the other. 

If we compare the motions of position sensors with visual sensors or image-acquired object motions, the relativistic reference can be expressed by Lorentz transformation.

\begin{equation}\label{eq:LorentzFaktor}
    \Gamma = \dfrac{1}{\sqrt{1-(\dfrac{v}{c})^{2}}} \in \mathbb{R}^{4} \hspace*{0.3cm} %with \hspace*{0.3cm} c^{2} = 1
\end{equation}

Generally, trajectory positions of $R,T$ can be calculated by integrating velocity over a discrete time period~\cite{MK.22}. By double-deriving the measured linear acceleration, we obtain the translational position.\\

\textit{First deviation:}
\vspace*{-0.3cm}
\begin{equation}\label{eq:VelocityResultV1}
   \Delta v= \int_{t_\mathrm{0}}^{t_\mathrm{n}}  \hspace{-0.1cm} a_\mathrm{cor} \hspace{0.1cm} \,dt + \Delta v_\mathrm{cor}
\end{equation}

\textit{Second deviation:}
\begin{equation}\label{eq:PositionResultV1}
    \Delta T = \bigl(\iint_{t_\mathrm{0}}^{t_\mathrm{n}} a_\mathrm{cor}^{i} \,dt \bigl)+ \Delta t \bigl(\int_{t_\mathrm{n}}^{(t_\mathrm{n}+\Delta)} \hspace*{-0.4cm} a_\mathrm{lin} \,dt\bigl) + \dfrac{\Delta t}{2} \Delta v_\mathrm{cor}
\end{equation}

The navigation depends on a referencable and accurate World Coordinate System (WCS) which is fixed on earth's frame position and derived to a relation for the corrected velocity \iffalse (Eq.~\ref{eq:VectorResultingVelocity})\fi \cite{kok18}. The resulting translation of Eq.~\ref{eq:PositionResultV1} comprises the translational changes of 4 spatial axes.

\begin{equation}\label{eq:HomogeneMatrix4DTranslation}
        \Delta T = \left(
        \begin{array}{c c c c c}
            1 & 0 & 0 & 0 & T_{\zeta}\\
            0 & -1 & 0 & 0 & T_\mathrm{x}\\ 
            0 & 0 & -1 & 0 & T_\mathrm{y}\\
            0 & 0 & 0 & -1 & T_\mathrm{z} \\ 
            0 & 0 & 0 & 0 & 1\\ 
        \end{array}
        \right)
\end{equation}

\vspace*{0.2cm}

Where we express $T_\mathrm{\zeta}$ as:

\begin{equation}\label{eq:Transformationskonstante}
    \mathcal{T}_{\zeta} =  \tau_\mathrm{\Gamma} \cdot (\sqrt{T_\mathrm{x}^{2} + T_\mathrm{y}^{2} + T_\mathrm{z}^{2}})    \\
\end{equation}

\vspace*{0.1cm}

The following term defines $\tau_\mathrm{\Gamma}$ as \textbf{Lorentz-Boost}:

\small{\begin{equation}\label{eq:LorentzBoost} \hspace*{-0.0cm} 
     \left(  \hspace*{-0.15cm}
        \begin{array}{c c c c}
            \Gamma & -\Gamma \dfrac{v_\mathrm{x}}{c} & -\Gamma \dfrac{v_\mathrm{y}}{c} & - \Gamma \dfrac{v_\mathrm{z}}{c} \\
            & & & \\
             -\Gamma \dfrac{v_\mathrm{x}}{c} & 1 + (\Gamma - 1) \dfrac{v_\mathrm{x}^{2}}{v^{2}}  & (\Gamma - 1) \dfrac{v_\mathrm{x} v_\mathrm{y}}{{v}^{2}} & (\Gamma - 1) \dfrac{v_\mathrm{x} v_\mathrm{z}}{{v}^{2}} \\ 
             & & & \\
            -\Gamma \dfrac{v_\mathrm{y}}{c} & (\Gamma - 1) \dfrac{v_\mathrm{y} v_\mathrm{x}}{{v}^{2}} & 1 + (\Gamma - 1) \dfrac{v_\mathrm{y}^{2}}{v^{2}} & (\Gamma - 1) \dfrac{v_\mathrm{y} v_\mathrm{z}}{{v}^{2}}\\
            & & & \\
            - \Gamma \dfrac{v_\mathrm{z}}{c} & (\Gamma - 1) \dfrac{v_\mathrm{z} v_\mathrm{x}}{{v}^{2}} & (\Gamma - 1) \dfrac{v_\mathrm{z} v_\mathrm{y}}{{v}^{2}} & 1 + (\Gamma - 1) \dfrac{v_\mathrm{z}^{2}}{v^{2}}  
        \end{array}    
        \\        
        \hspace*{-0.15cm} \right)  \hspace*{0.4cm}  \vspace*{0.2cm}
\end{equation}}
\\
\begin{figure*}[ht!]
    \centering

\tikzset{every picture/.style={line width=0.75pt}} %set default line width to 0.75pt        

\begin{tikzpicture}[x=0.75pt,y=0.75pt,yscale=-0.93,xscale=0.93]
%uncomment if require: \path (0,7037); %set diagram left start at 0, and has height of 7037

%Shape: Cube [id:dp3118077994853039] 
\draw  [color={rgb, 255:red, 255; green, 255; blue, 255 }  ,draw opacity=1 ][fill={rgb, 255:red, 74; green, 74; blue, 74 }  ,fill opacity=0.91 ] (426.6,6804.58) -- (407.91,6786.22) -- (407.83,6777.18) -- (440.47,6776.89) -- (459.16,6795.26) -- (459.25,6804.29) -- cycle ; \draw  [color={rgb, 255:red, 255; green, 255; blue, 255 }  ,draw opacity=1 ] (407.83,6777.18) -- (426.52,6795.55) -- (426.6,6804.58) ; \draw  [color={rgb, 255:red, 255; green, 255; blue, 255 }  ,draw opacity=1 ] (426.52,6795.55) -- (459.16,6795.26) ;
%Shape: Cube [id:dp3355057901278622] 
\draw  [color={rgb, 255:red, 255; green, 255; blue, 255 }  ,draw opacity=1 ][fill={rgb, 255:red, 74; green, 74; blue, 74 }  ,fill opacity=0.91 ] (532.6,6803.58) -- (513.91,6785.22) -- (513.83,6776.18) -- (546.47,6775.89) -- (565.16,6794.26) -- (565.25,6803.29) -- cycle ; \draw  [color={rgb, 255:red, 255; green, 255; blue, 255 }  ,draw opacity=1 ] (513.83,6776.18) -- (532.52,6794.55) -- (532.6,6803.58) ; \draw  [color={rgb, 255:red, 255; green, 255; blue, 255 }  ,draw opacity=1 ] (532.52,6794.55) -- (565.16,6794.26) ;
%Shape: Cube [id:dp6537798951488907] 
\draw  [color={rgb, 255:red, 255; green, 255; blue, 255 }  ,draw opacity=1 ][fill={rgb, 255:red, 2; green, 157; blue, 196 }  ,fill opacity=0.58 ] (202.68,6685.61) -- (183.99,6667.25) -- (183.91,6658.22) -- (216.55,6657.93) -- (235.25,6676.29) -- (235.33,6685.32) -- cycle ; \draw  [color={rgb, 255:red, 255; green, 255; blue, 255 }  ,draw opacity=1 ] (183.91,6658.22) -- (202.6,6676.58) -- (202.68,6685.61) ; \draw  [color={rgb, 255:red, 255; green, 255; blue, 255 }  ,draw opacity=1 ] (202.6,6676.58) -- (235.25,6676.29) ;
%Shape: Cube [id:dp7586479069572201] 
\draw  [color={rgb, 255:red, 255; green, 255; blue, 255 }  ,draw opacity=1 ][fill={rgb, 255:red, 74; green, 74; blue, 74 }  ,fill opacity=0.91 ] (571.6,6735.58) -- (552.91,6717.22) -- (552.83,6708.18) -- (585.47,6707.89) -- (604.16,6726.26) -- (604.25,6735.29) -- cycle ; \draw  [color={rgb, 255:red, 255; green, 255; blue, 255 }  ,draw opacity=1 ] (552.83,6708.18) -- (571.52,6726.55) -- (571.6,6735.58) ; \draw  [color={rgb, 255:red, 255; green, 255; blue, 255 }  ,draw opacity=1 ] (571.52,6726.55) -- (604.16,6726.26) ;
%Shape: Cube [id:dp2999376252412673] 
\draw  [color={rgb, 255:red, 255; green, 255; blue, 255 }  ,draw opacity=1 ][fill={rgb, 255:red, 74; green, 74; blue, 74 }  ,fill opacity=0.91 ] (451.37,6717.48) -- (398.44,6665.48) -- (398.34,6655.08) -- (431.33,6654.78) -- (484.26,6706.78) -- (484.36,6717.19) -- cycle ; \draw  [color={rgb, 255:red, 255; green, 255; blue, 255 }  ,draw opacity=1 ] (398.34,6655.08) -- (451.28,6707.07) -- (451.37,6717.48) ; \draw  [color={rgb, 255:red, 255; green, 255; blue, 255 }  ,draw opacity=1 ] (451.28,6707.07) -- (484.26,6706.78) ;
%Straight Lines [id:da4103893126609517] 
\draw    (135.36,6667.25) -- (185.99,6667.25) ;
\draw [shift={(188.99,6667.25)}, rotate = 180] [fill={rgb, 255:red, 0; green, 0; blue, 0 }  ][line width=0.08]  [draw opacity=0] (8.04,-3.86) -- (0,0) -- (8.04,3.86) -- (5.34,0) -- cycle    ;
%Straight Lines [id:da3499217157286074] 
\draw    (141.36,6675.25) -- (193.99,6675.25) ;
\draw [shift={(196.99,6675.25)}, rotate = 180] [fill={rgb, 255:red, 0; green, 0; blue, 0 }  ][line width=0.08]  [draw opacity=0] (8.04,-3.86) -- (0,0) -- (8.04,3.86) -- (5.34,0) -- cycle    ;
%Straight Lines [id:da2968295500448066] 
\draw    (231.36,6671.25) -- (263.36,6671.25) ;
%Straight Lines [id:da5938495386725908] 
\draw  [dash pattern={on 4.5pt off 4.5pt}]  (209.36,6631.19) -- (209.36,6663.19) ;
\draw [shift={(209.36,6666.19)}, rotate = 270] [fill={rgb, 255:red, 0; green, 0; blue, 0 }  ][line width=0.08]  [draw opacity=0] (8.04,-3.86) -- (0,0) -- (8.04,3.86) -- (5.34,0) -- cycle    ;
%Shape: Diamond [id:dp9039526586507329] 
\draw  [color={rgb, 255:red, 255; green, 255; blue, 255 }  ,draw opacity=1 ][fill={rgb, 255:red, 9; green, 60; blue, 129 }  ,fill opacity=1 ] (267.86,6667.11) -- (272.71,6671.05) -- (268.21,6675.39) -- (263.36,6671.45) -- cycle ;
%Shape: Cube [id:dp9142012021556911] 
\draw  [color={rgb, 255:red, 255; green, 255; blue, 255 }  ,draw opacity=1 ][fill={rgb, 255:red, 74; green, 74; blue, 74 }  ,fill opacity=0.91 ] (316.68,6685.61) -- (297.99,6667.25) -- (297.91,6658.22) -- (330.55,6657.93) -- (349.25,6676.29) -- (349.33,6685.32) -- cycle ; \draw  [color={rgb, 255:red, 255; green, 255; blue, 255 }  ,draw opacity=1 ] (297.91,6658.22) -- (316.6,6676.58) -- (316.68,6685.61) ; \draw  [color={rgb, 255:red, 255; green, 255; blue, 255 }  ,draw opacity=1 ] (316.6,6676.58) -- (349.25,6676.29) ;
%Straight Lines [id:da7727957946557489] 
\draw    (272.71,6671.25) -- (301.71,6671.25) ;
\draw [shift={(304.71,6671.25)}, rotate = 180] [fill={rgb, 255:red, 0; green, 0; blue, 0 }  ][line width=0.08]  [draw opacity=0] (8.04,-3.86) -- (0,0) -- (8.04,3.86) -- (5.34,0) -- cycle    ;
%Straight Lines [id:da7208330985425422] 
\draw  [dash pattern={on 4.5pt off 4.5pt}]  (321.36,6631.19) -- (321.36,6663.19) ;
\draw [shift={(321.36,6666.19)}, rotate = 270] [fill={rgb, 255:red, 0; green, 0; blue, 0 }  ][line width=0.08]  [draw opacity=0] (8.04,-3.86) -- (0,0) -- (8.04,3.86) -- (5.34,0) -- cycle    ;
%Straight Lines [id:da1484135860631708] 
\draw    (345.71,6670.25) -- (399.36,6670.25) ;
\draw [shift={(402.36,6670.25)}, rotate = 180] [fill={rgb, 255:red, 0; green, 0; blue, 0 }  ][line width=0.08]  [draw opacity=0] (8.04,-3.86) -- (0,0) -- (8.04,3.86) -- (5.34,0) -- cycle    ;
%Straight Lines [id:da8774748108392074] 
\draw    (270.36,6675.62) -- (302.36,6709.19) -- (439.36,6709.19) ;
\draw [shift={(442.36,6709.19)}, rotate = 180] [fill={rgb, 255:red, 0; green, 0; blue, 0 }  ][line width=0.08]  [draw opacity=0] (8.04,-3.86) -- (0,0) -- (8.04,3.86) -- (5.34,0) -- cycle    ;
%Straight Lines [id:da42658541633172975] 
\draw  [dash pattern={on 4.5pt off 4.5pt}]  (438.36,6629.19) -- (438.36,6672.19) ;
\draw [shift={(438.36,6675.19)}, rotate = 270] [fill={rgb, 255:red, 0; green, 0; blue, 0 }  ][line width=0.08]  [draw opacity=0] (8.04,-3.86) -- (0,0) -- (8.04,3.86) -- (5.34,0) -- cycle    ;
%Straight Lines [id:da3848620049644187] 
\draw    (463.36,6684.25) -- (542.36,6684.25) -- (563.24,6705.13) ;
\draw [shift={(565.36,6707.25)}, rotate = 225] [fill={rgb, 255:red, 0; green, 0; blue, 0 }  ][line width=0.08]  [draw opacity=0] (8.04,-3.86) -- (0,0) -- (8.04,3.86) -- (5.34,0) -- cycle    ;
%Straight Lines [id:da4138718576603071] 
\draw    (455.36,6789.25) -- (518.99,6789.25) ;
\draw [shift={(521.99,6789.25)}, rotate = 180] [fill={rgb, 255:red, 0; green, 0; blue, 0 }  ][line width=0.08]  [draw opacity=0] (8.04,-3.86) -- (0,0) -- (8.04,3.86) -- (5.34,0) -- cycle    ;
%Straight Lines [id:da7056888095276006] 
\draw  [dash pattern={on 4.5pt off 4.5pt}]  (540.36,6751.19) -- (540.36,6783.19) ;
\draw [shift={(540.36,6786.19)}, rotate = 270] [fill={rgb, 255:red, 0; green, 0; blue, 0 }  ][line width=0.08]  [draw opacity=0] (8.04,-3.86) -- (0,0) -- (8.04,3.86) -- (5.34,0) -- cycle    ;
%Straight Lines [id:da23019082136146274] 
\draw [color={rgb, 255:red, 255; green, 255; blue, 255 }  ,draw opacity=1 ] [dash pattern={on 4.5pt off 4.5pt}]  (343.67,6709.19) -- (371.86,6709.19) ;
%Straight Lines [id:da18209094228094602] 
\draw  [dash pattern={on 4.5pt off 4.5pt}]  (433.36,6749.19) -- (433.36,6781.19) ;
\draw [shift={(433.36,6784.19)}, rotate = 270] [fill={rgb, 255:red, 0; green, 0; blue, 0 }  ][line width=0.08]  [draw opacity=0] (8.04,-3.86) -- (0,0) -- (8.04,3.86) -- (5.34,0) -- cycle    ;
%Shape: Cube [id:dp16235727314971005] 
\draw  [color={rgb, 255:red, 255; green, 255; blue, 255 }  ,draw opacity=1 ][fill={rgb, 255:red, 2; green, 157; blue, 196 }  ,fill opacity=0.58 ] (206.6,6761.58) -- (187.91,6743.22) -- (187.83,6734.18) -- (220.47,6733.89) -- (239.16,6752.26) -- (239.25,6761.29) -- cycle ; \draw  [color={rgb, 255:red, 255; green, 255; blue, 255 }  ,draw opacity=1 ] (187.83,6734.18) -- (206.52,6752.55) -- (206.6,6761.58) ; \draw  [color={rgb, 255:red, 255; green, 255; blue, 255 }  ,draw opacity=1 ] (206.52,6752.55) -- (239.16,6752.26) ;
%Straight Lines [id:da17189542826490856] 
\draw  [dash pattern={on 4.5pt off 4.5pt}]  (212.36,6707.19) -- (212.36,6739.19) ;
\draw [shift={(212.36,6742.19)}, rotate = 270] [fill={rgb, 255:red, 0; green, 0; blue, 0 }  ][line width=0.08]  [draw opacity=0] (8.04,-3.86) -- (0,0) -- (8.04,3.86) -- (5.34,0) -- cycle    ;
%Straight Lines [id:da40602508655730474] 
\draw    (234.36,6747.33) -- (395.36,6747.33) ;
%Shape: Cube [id:dp18195915292227083] 
\draw  [color={rgb, 255:red, 255; green, 255; blue, 255 }  ,draw opacity=1 ][fill={rgb, 255:red, 2; green, 157; blue, 196 }  ,fill opacity=0.58 ] (132.6,6762.58) -- (113.91,6744.22) -- (113.83,6735.18) -- (146.47,6734.89) -- (165.16,6753.26) -- (165.25,6762.29) -- cycle ; \draw  [color={rgb, 255:red, 255; green, 255; blue, 255 }  ,draw opacity=1 ] (113.83,6735.18) -- (132.52,6753.55) -- (132.6,6762.58) ; \draw  [color={rgb, 255:red, 255; green, 255; blue, 255 }  ,draw opacity=1 ] (132.52,6753.55) -- (165.16,6753.26) ;

%Straight Lines [id:da10594475837816342] 
\draw    (157.36,6747.33) -- (191,6747.33) ;
\draw [shift={(194,6747.33)}, rotate = 180] [fill={rgb, 255:red, 0; green, 0; blue, 0 }  ][line width=0.08]  [draw opacity=0] (8.04,-3.86) -- (0,0) -- (8.04,3.86) -- (5.34,0) -- cycle    ;
%Straight Lines [id:da3418619765522808] 
\draw  [dash pattern={on 4.5pt off 4.5pt}]  (137.36,6707.19) -- (137.36,6739.19) ;
\draw [shift={(137.36,6742.19)}, rotate = 270] [fill={rgb, 255:red, 0; green, 0; blue, 0 }  ][line width=0.08]  [draw opacity=0] (8.04,-3.86) -- (0,0) -- (8.04,3.86) -- (5.34,0) -- cycle    ;
%Straight Lines [id:da1396310767093898] 
\draw    (66.36,6746.25) -- (116.99,6746.25) ;
\draw [shift={(119.99,6746.25)}, rotate = 180] [fill={rgb, 255:red, 0; green, 0; blue, 0 }  ][line width=0.08]  [draw opacity=0] (8.04,-3.86) -- (0,0) -- (8.04,3.86) -- (5.34,0) -- cycle    ;
%Shape: Diamond [id:dp06696406832781987] 
\draw  [color={rgb, 255:red, 255; green, 255; blue, 255 }  ,draw opacity=1 ][fill={rgb, 255:red, 9; green, 60; blue, 129 }  ,fill opacity=1 ] (394.22,6742.45) -- (400.06,6744.68) -- (397.14,6750.21) -- (391.3,6747.98) -- cycle ;
%Straight Lines [id:da17134994415596494] 
\draw    (393.36,6743.62) -- (333.22,6682.75) ;
\draw [shift={(331.11,6680.62)}, rotate = 45.34] [fill={rgb, 255:red, 0; green, 0; blue, 0 }  ][line width=0.08]  [draw opacity=0] (8.04,-3.86) -- (0,0) -- (8.04,3.86) -- (5.34,0) -- cycle    ;
%Straight Lines [id:da43908496313866485] 
\draw    (398.36,6750.62) -- (421.2,6773.77) ;
\draw [shift={(423.3,6775.9)}, rotate = 225.39] [fill={rgb, 255:red, 0; green, 0; blue, 0 }  ][line width=0.08]  [draw opacity=0] (8.04,-3.86) -- (0,0) -- (8.04,3.86) -- (5.34,0) -- cycle    ;
%Straight Lines [id:da4528363145751899] 
\draw    (411.36,6817.25) -- (564.99,6817.25) -- (549.47,6801.74) ;
\draw [shift={(547.35,6799.62)}, rotate = 45] [fill={rgb, 255:red, 0; green, 0; blue, 0 }  ][line width=0.08]  [draw opacity=0] (8.04,-3.86) -- (0,0) -- (8.04,3.86) -- (5.34,0) -- cycle    ;
%Straight Lines [id:da45691127814369215] 
\draw    (66.36,6746.25) -- (135.36,6667.25) ;
%Shape: Diamond [id:dp946761287407007] 
\draw  [color={rgb, 255:red, 255; green, 255; blue, 255 }  ,draw opacity=1 ][fill={rgb, 255:red, 9; green, 60; blue, 129 }  ,fill opacity=1 ] (90.68,6715.19) -- (95.36,6719.33) -- (90.68,6723.47) -- (86,6719.33) -- cycle ;
%Straight Lines [id:da42953663988540103] 
\draw    (57.36,6719.33) -- (83,6719.33) ;
\draw [shift={(86,6719.33)}, rotate = 180] [fill={rgb, 255:red, 0; green, 0; blue, 0 }  ][line width=0.08]  [draw opacity=0] (8.04,-3.86) -- (0,0) -- (8.04,3.86) -- (5.34,0) -- cycle    ;
%Straight Lines [id:da4696182609370163] 
\draw    (561.36,6787.25) -- (642.36,6787.25) -- (589.48,6734.37) ;
\draw [shift={(587.36,6732.25)}, rotate = 45] [fill={rgb, 255:red, 0; green, 0; blue, 0 }  ][line width=0.08]  [draw opacity=0] (8.04,-3.86) -- (0,0) -- (8.04,3.86) -- (5.34,0) -- cycle    ;
%Straight Lines [id:da5428253536543779] 
\draw    (597.36,6718.25) -- (649.99,6718.25) ;
\draw [shift={(652.99,6718.25)}, rotate = 180] [fill={rgb, 255:red, 0; green, 0; blue, 0 }  ][line width=0.08]  [draw opacity=0] (8.04,-3.86) -- (0,0) -- (8.04,3.86) -- (5.34,0) -- cycle    ;
%Straight Lines [id:da015002236738221653] 
\draw  [dash pattern={on 4.5pt off 4.5pt}]  (579.36,6674.62) -- (579.36,6711.19) ;
\draw [shift={(579.36,6714.19)}, rotate = 270] [fill={rgb, 255:red, 0; green, 0; blue, 0 }  ][line width=0.08]  [draw opacity=0] (8.04,-3.86) -- (0,0) -- (8.04,3.86) -- (5.34,0) -- cycle    ;

%Shape: Cube [id:dp5159629755427371] 
\draw  [color={rgb, 255:red, 255; green, 255; blue, 255 }  ,draw opacity=1 ][fill={rgb, 255:red, 2; green, 157; blue, 196 }  ,fill opacity=0.58 ] (384.6,6830.58) -- (365.91,6812.22) -- (365.83,6803.18) -- (398.47,6802.89) -- (417.16,6821.26) -- (417.25,6830.29) -- cycle ; \draw  [color={rgb, 255:red, 255; green, 255; blue, 255 }  ,draw opacity=1 ] (365.83,6803.18) -- (384.52,6821.55) -- (384.6,6830.58) ; \draw  [color={rgb, 255:red, 255; green, 255; blue, 255 }  ,draw opacity=1 ] (384.52,6821.55) -- (417.16,6821.26) ;

%Straight Lines [id:da2907117328886155] 
\draw  [dash pattern={on 4.5pt off 4.5pt}]  (389.36,6773.33) -- (389.36,6807.19) ;
\draw [shift={(389.36,6810.19)}, rotate = 270] [fill={rgb, 255:red, 0; green, 0; blue, 0 }  ][line width=0.08]  [draw opacity=0] (8.04,-3.86) -- (0,0) -- (8.04,3.86) -- (5.34,0) -- cycle    ;

%Straight Lines [id:da6024938057563882] 
\draw    (344.36,6818.33) -- (370,6818.33) ;
\draw [shift={(373,6818.33)}, rotate = 180] [fill={rgb, 255:red, 0; green, 0; blue, 0 }  ][line width=0.08]  [draw opacity=0] (8.04,-3.86) -- (0,0) -- (8.04,3.86) -- (5.34,0) -- cycle    ;

%Straight Lines [id:da030248794399911905] 
\draw    (383.36,6802.33) -- (372.36,6791.33) -- (411.36,6791.33) ;
\draw [shift={(414.36,6791.33)}, rotate = 180] [fill={rgb, 255:red, 0; green, 0; blue, 0 }  ][line width=0.08]  [draw opacity=0] (8.04,-3.86) -- (0,0) -- (8.04,3.86) -- (5.34,0) -- cycle    ;

%Shape: Cube [id:dp9535744082520248] 
\draw  [color={rgb, 255:red, 255; green, 255; blue, 255 }  ,draw opacity=1 ][fill={rgb, 255:red, 2; green, 157; blue, 196 }  ,fill opacity=0.58 ] (85.6,6823.58) -- (66.91,6805.22) -- (66.83,6796.18) -- (99.47,6795.89) -- (118.16,6814.26) -- (118.25,6823.29) -- cycle ; \draw  [color={rgb, 255:red, 255; green, 255; blue, 255 }  ,draw opacity=1 ] (66.83,6796.18) -- (85.52,6814.55) -- (85.6,6823.58) ; \draw  [color={rgb, 255:red, 255; green, 255; blue, 255 }  ,draw opacity=1 ] (85.52,6814.55) -- (118.16,6814.26) ;
%Shape: Cube [id:dp15023028289260942] 
\draw  [color={rgb, 255:red, 255; green, 255; blue, 255 }  ,draw opacity=1 ][fill={rgb, 255:red, 74; green, 74; blue, 74 }  ,fill opacity=0.84 ] (84.6,6807.58) -- (65.91,6789.22) -- (65.83,6780.18) -- (98.47,6779.89) -- (117.16,6798.26) -- (117.25,6807.29) -- cycle ; \draw  [color={rgb, 255:red, 255; green, 255; blue, 255 }  ,draw opacity=1 ] (65.83,6780.18) -- (84.52,6798.55) -- (84.6,6807.58) ; \draw  [color={rgb, 255:red, 255; green, 255; blue, 255 }  ,draw opacity=1 ] (84.52,6798.55) -- (117.16,6798.26) ;

% Text Node
\draw (123,6790) node [anchor=north west][inner sep=0.75pt]   [align=left] {$V^{*} \in \mathbb{R}^{4}$};
% Text Node
\draw (123.25,6808) node [anchor=north west][inner sep=0.75pt]   [align=left] {$V^{} \in \mathbb{R}^{3}$};

% Text Node
\draw (340,6772) node [anchor=north west][inner sep=0.75pt]   [align=left] {(Eq.\ref{eq:Angles})};

% Text Node
%\draw (339,6770.9) node [anchor=north west][inner sep=0.75pt]   [align=left] {[$\psi,\theta,\phi$]};

% Text Node
\draw (355,6755) node [anchor=north west][inner sep=0.75pt]   [align=left] {\begin{minipage}[lt]{36.73pt}\setlength\topsep{0pt}
\begin{center}
{Angles}
\end{center}

\end{minipage}};

% Text Node
\draw (160,6678) node [anchor=north west][inner sep=0.75pt]   [align=left] {$B$};
% Text Node
\draw (164,6595) node [anchor=north west][inner sep=0.75pt]   [align=left] {\begin{minipage}[lt]{59.97pt}\setlength\topsep{0pt}
\begin{center}
{3D }\\{Translation }
\end{center}

\end{minipage}};
% Text Node
\draw (230,6650.9) node [anchor=north west][inner sep=0.75pt]   [align=left] {$\nabla T \in \mathbb{R}^{3}$};
% Text Node
\draw (212,6635) node [anchor=north west][inner sep=0.75pt]   [align=left] {(Eq.\ref{eq:PositionResultV1})};
% Text Node
\draw (323,6635) node [anchor=north west][inner sep=0.75pt]   [align=left] {(Eq.\ref{eq:Transformationskonstante})};
% Text Node
\draw (274,6600) node [anchor=north west][inner sep=0.75pt]   [align=left] {\begin{minipage}[lt]{59.97pt}\setlength\topsep{0pt}
\begin{center}
{Temporal }\\{Translation }
\end{center}

\end{minipage}};
% Text Node
\draw (362,6652) node [anchor=north west][inner sep=0.75pt]   [align=left] {$T_{\zeta}$};
% Text Node
%\draw (365,6595) node [anchor=north west][inner sep=0.75pt]   [align=left] {$\nabla T \in \mathbb{R}^{3}$};
% Text Node
\draw (440,6635) node [anchor=north west][inner sep=0.75pt]   [align=left] {(Eq.\ref{eq:HomogeneMatrix4DTranslation})};
% Text Node
\draw (393,6595) node [anchor=north west][inner sep=0.75pt]   [align=left] {\begin{minipage}[lt]{59.97pt}\setlength\topsep{0pt}
\begin{center}
{4D }\\{Translation }
\end{center}

\end{minipage}};
% Text Node
\draw (470,6663.9) node [anchor=north west][inner sep=0.75pt]   [align=left] {$\nabla T \in \mathbb{R}^{4}$};
% Text Node
\draw (450,6768.9) node [anchor=north west][inner sep=0.75pt]   [align=left] {[$\psi^{'},\theta^{'},\phi^{'}$]};
% Text Node
\draw (505,6718.9) node [anchor=north west][inner sep=0.75pt]   [align=left] {\begin{minipage}[lt]{46.91pt}\setlength\topsep{0pt}
\begin{center}
{4D}\\ {Rotation}
\end{center}

\end{minipage}};
% Text Node
\draw (541.68,6755) node [anchor=north west][inner sep=0.75pt]   [align=left] {(Eq.\ref{eq:lorentzrotation} )};
% Text Node
\draw (398,6718) node [anchor=north west][inner sep=0.75pt]   [align=left] {\begin{minipage}[lt]{80.71pt}\setlength\topsep{0pt}
\begin{center}
{Temporal}\\{Shifted-Angles}
\end{center}

\end{minipage}};
% Text Node
\draw (435,6755) node [anchor=north west][inner sep=0.75pt]   [align=left] {(Eq.\ref{eq:WinkelTransformation})};
% Text Node
\draw (160,6690.9) node [anchor=north west][inner sep=0.75pt]   [align=left] {\begin{minipage}[lt]{71.85pt}\setlength\topsep{0pt}
\begin{center}
{Lorentz}
\end{center}

\end{minipage}};
% Text Node
\draw (214,6710) node [anchor=north west][inner sep=0.75pt]   [align=left] {(Eq.\ref{eq:LorentzBoost})};
% Text Node
\draw (320,6688) node [anchor=north west][inner sep=0.75pt]   [align=left] {$\Gamma_{\tau}$};
% Text Node
\draw (109,6690.9) node [anchor=north west][inner sep=0.75pt]   [align=left] {\begin{minipage}[lt]{41.28pt}\setlength\topsep{0pt}
\begin{center}
{Velocity}
\end{center}

\end{minipage}};
% Text Node
\draw (139,6710) node [anchor=north west][inner sep=0.75pt]   [align=left] {(Eq.\ref{eq:VelocityResultV1})};
% Text Node
\draw (64,6706) node [anchor=north west][inner sep=0.75pt]   [align=left] {$a$};
% Text Node
\draw (268.25,6730) node [anchor=north west][inner sep=0.75pt]   [align=left] {$\Gamma_{\tau}$};
% Text Node
\draw (348,6805) node [anchor=north west][inner sep=0.75pt]   [align=left] {$\omega$};
% Text Node
\draw (460,6800) node [anchor=north west][inner sep=0.75pt]   [align=left] {[$\psi,\theta,\phi$]};

% Text Node
\draw (160,6730) node [anchor=north west][inner sep=0.75pt]   [align=left] {$\nabla v$};
% Text Node
\draw (565,6770) node [anchor=north west][inner sep=0.75pt]   [align=left] {$\nabla R \in \mathbb{R}^{4}$};
% Text Node
\draw (550,6643) node [anchor=north west][inner sep=0.75pt]   [align=left] {\begin{minipage}[lt]{42.96pt}\setlength\topsep{0pt}
\begin{center}
{4D }\\{Position}
\end{center}

\end{minipage}};
% Text Node
\draw (582,6680) node [anchor=north west][inner sep=0.75pt]   [align=left] {(Eq.\ref{eq:Spatio-Temoral-Motion})};
% Text Node
\draw (614,6698.9) node [anchor=north west][inner sep=0.75pt]   [align=left] {$\chi^{\mu}$};
\end{tikzpicture}

    \caption{The illustration shows the design of sensor estimated 4D-positioning. The respective rotation and translation can be calculated from IMU-specific acceleration, angular velocities, and magnetic fields. By applying the Lorentz factor, the temporal translation and rotations can be determined. The resulting 4D translation and rotation can be expressed as a 4D position in the form of the four-vector $\chi$. }
    \label{fig:DesignSensorEstimatedPositioning}
\end{figure*}
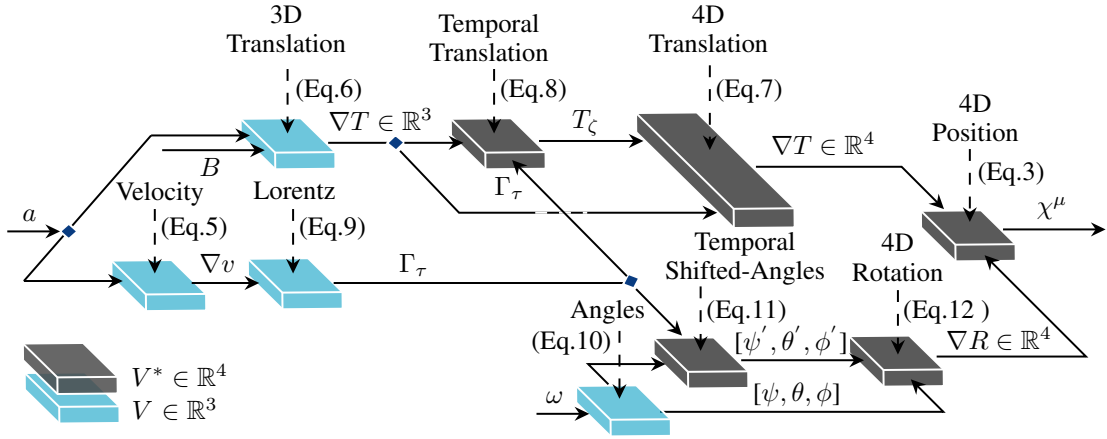
\vspace*{-0.3cm}

The angular displacement can be estimated by the sensor-detected angular velocity:

\begin{equation}\label{eq:Angles}
  R \rightarrow \varphi [^{\circ}]:  \hspace{0.2cm}  \Delta \varphi_\mathrm{(t_\mathrm{})} =     \int  \hspace*{-0.0cm} \omega_\mathrm{} \,dt \hspace*{0.784cm}
\end{equation}

Where $\psi^{'},\theta^{'}$ and $\phi^{'}$ are defined as:

\begin{equation}\label{eq:WinkelTransformation}
    [\psi^{'},\theta^{'},\phi^{'} ] = [\psi,\theta,\phi ] (1 + \tau_{\Gamma})
\end{equation}

\vspace*{0.1cm}
Besides translation, the spatial alignment can be described in terms of the 6 rotational equations.

\begin{table}[ht!]
    \centering
    \begin{tabular}{c c}
  $R_{x,y}:$   & $R_{x,z}:$  \\ 
  & \vspace*{-0.3cm} \\
          \small{ $\left(
        \begin{array}{c c c c}
            cos\psi & -sin\psi & 0 & 0\\
            sin\psi & cos\psi & 0 & 0\\ 
            0 & 0 & 1 & 0\\ 
             0 & 0 & 0 & 1 \\
        \end{array}
        \right)$  } & \hspace*{-0.4cm}  \small{  $\left(
    \begin{array}{c c c c}
        cos\theta & 0 & sin\theta & 0\\
        0 & 1 & 0 & 0\\ 
        -sin\theta & 0 & cos\theta & 0\\
        0 & 0 & 0 & 1 \\
    \end{array}
    \right)$ } \\
    % & \vspace*{-0.3cm} \\ 
   % & \\\hline
           & \vspace*{-0.3cm} \\ 
           $R_{y,z}:$ & $R_{\zeta,z}:$ \\
 & \vspace*{-0.3cm} \\ 
        \small{    $\left(
        \begin{array}{c c c c}
            1 & 0 & 0& 0\\
            0 & cos\phi & -sin\phi & 0\\ 
            0 & sin\phi & cos\phi & 0\\ 
             0 & 0 & 0 & 1 \\
        \end{array}
        \right)  $ } & \hspace*{-0.4cm} \small{ $ \left(
        \begin{array}{c c c c}
            1 & 0 & 0 & 0\\
            0 & 1 & 0 & 0\\ 
            0 & 0 & cos\psi' & sin\psi'\\ 
             0 & 0 & -sin\psi' & cos\psi' \\
        \end{array}
      \hspace*{-0.2cm}  \right)$ } \\
 %         & \\\hline
 & \vspace*{-0.3cm} \\ 
          $R_{\zeta,y}:$ & $R_{\zeta,x}$ \\
 & \vspace*{-0.3cm} \\ 
\small{ $\left(
        \begin{array}{c c c c}
            1 & 0 & 0 & 0\\
            0 & cos\theta' & 0 & sin\theta'\\ 
            0 & 0 & 1 & 0\\ 
             0 & -sin\theta' & 0 & cos\theta' \\
        \end{array}
        \right)$ } &  \hspace*{-0.4cm} \small{ $\left(
        \begin{array}{c c c c}
            cos\phi' & 0 & 0 &  -sin\phi'\\
            0 & 1 & 0 & 0\\ 
            0 & 0 & 1 & 0\\ 
            sin\phi' & 0 & 0 & cos\phi' \\
        \end{array}
       \hspace*{-0.15cm}  \right )$ }  
       \vspace*{-1cm}
    \end{tabular}
  %  \caption{Caption}
  %  \label{tab:my_label}
\end{table}
%\vspace*{-0.4cm}
\begin{equation}\label{eq:lorentzrotation} 
\end{equation}

\normalsize
%\subsection{Formulation of 4D-Navigated Motion}

\vspace*{0.0cm}
The described definitions form the fundamental criteria for the relativistic processing of sensor information. This means that acquired sensor variables such as accelerations or forces can be related to each other in terms of time to estimate positions or describe relative movements of different systems. This variant improves the navigation of robotic systems through the temporal correlation and contextualization of sensor data and in an environment.

\subsection{Design of Sensor based 4D-Navigation}

The design of 4D-based navigation shall be introduced employing the presented equations. We restrict our focus to the accelerations $a$, angular velocities $\omega$ as well as magnetic fields $B$ of sensor-integrated Inertial Measurement Unit (IMU), and the associated image data. 

Fig.~\ref{fig:DesignSensorEstimatedPositioning} introduces the design of spatio-temporal position estimation for robot navigation in a common reference system using different sensor sizes. We calculate the velocity $\nabla v$ in the x, y, and z directions with the acquired acceleration by sensors. 
With $\nabla v$ we calculate the Lorentz boost $\Gamma_{\tau}$ which refers to the data of previous time $\tau_{n-1}$. The temporal translation $T_\mathrm{\zeta}$ can be derived from $\Gamma_{\tau}$ and the previously calculated translation $\nabla T \in \mathbb{R}^{3}$. In reverse, the angles around the temporal axes [$\psi^{'},\theta^{'},\phi^{'}$] can be calculated using $\Gamma_{\tau}$ and $\omega$ in order to define the 4D rotations. The resulting 4D translation and rotation can be expressed as a 4D position in form of the four-vector $\chi$. 

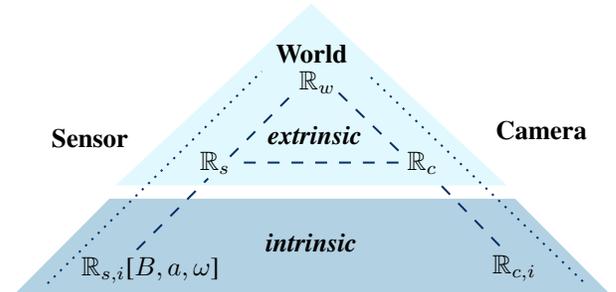
\begin{figure}[ht!]
    \centering

\tikzset{every picture/.style={line width=0.75pt}} %set default line width to 0.75pt        

\begin{tikzpicture}[x=0.75pt,y=0.75pt,yscale=-1,xscale=1]
%uncomment if require: \path (0,7367); %set diagram left start at 0, and has height of 7367

%Shape: Trapezoid [id:dp4326382638554924] 
\draw  [color={rgb, 255:red, 255; green, 255; blue, 255 }  ,draw opacity=1 ][fill={rgb, 255:red, 15; green, 113; blue, 167 }  ,fill opacity=0.32 ] (228.36,7282.62) -- (277.36,7233.33) -- (482.36,7233.33) -- (531.36,7282.62) -- cycle ;
%Shape: Triangle [id:dp13729291927221787] 
\draw  [color={rgb, 255:red, 255; green, 255; blue, 255 }  ,draw opacity=1 ][fill={rgb, 255:red, 8; green, 211; blue, 248 }  ,fill opacity=0.12 ] (379.36,7134.55) -- (479.36,7227.62) -- (279.36,7227.62) -- cycle ;

%Straight Lines [id:da63911705889711] 
\draw [color={rgb, 255:red, 8; green, 46; blue, 103 }  ,draw opacity=1 ][line width=0.75]  [dash pattern={on 4.5pt off 4.5pt}]  (338.36,7212.62) -- (369.36,7181.62) ;
%Straight Lines [id:da35712568998453165] 
\draw [color={rgb, 255:red, 8; green, 46; blue, 103 }  ,draw opacity=1 ][line width=0.75]  [dash pattern={on 4.5pt off 4.5pt}]  (291.36,7259.62) -- (327.36,7223.62) ;
%Straight Lines [id:da37193184779577826] 
\draw [color={rgb, 255:red, 8; green, 46; blue, 103 }  ,draw opacity=1 ][line width=0.75]  [dash pattern={on 4.5pt off 4.5pt}]  (423.36,7209.62) -- (391.36,7178.62) ;
%Straight Lines [id:da13933184654063058] 
\draw [color={rgb, 255:red, 8; green, 46; blue, 103 }  ,draw opacity=1 ][line width=0.75]  [dash pattern={on 4.5pt off 4.5pt}]  (474.36,7257.62) -- (443.36,7225.62) ;
%Straight Lines [id:da18914344801448402] 
\draw [color={rgb, 255:red, 8; green, 46; blue, 103 }  ,draw opacity=1 ][line width=0.75]  [dash pattern={on 4.5pt off 4.5pt}]  (345.36,7215.62) -- (423.36,7215.62) ;
%Straight Lines [id:da3729140913833038] 
\draw [color={rgb, 255:red, 8; green, 46; blue, 103 }  ,draw opacity=1 ][line width=0.75]  [dash pattern={on 0.84pt off 2.51pt}]  (246.36,7276.62) -- (354.36,7169.62) ;
%Straight Lines [id:da0686627856261659] 
\draw [color={rgb, 255:red, 8; green, 46; blue, 103 }  ,draw opacity=1 ][line width=0.75]  [dash pattern={on 0.84pt off 2.51pt}]  (516.36,7275.62) -- (407.36,7170.62) ;

% Text Node
\draw (360,7155) node [anchor=north west][inner sep=0.75pt]   [align=left] {\textbf{World}};
% Text Node
\draw (355,7250) node [anchor=north west][inner sep=0.75pt]   [align=left] {\textit{\textbf{intrinsic}}};
% Text Node
\draw (262,7262) node [anchor=north west][inner sep=0.75pt]   [align=left] {\textbf{$\mathbb{R}_{s,i}$}[$B,a,\omega$]};
% Text Node
\draw (466,7260.9) node [anchor=north west][inner sep=0.75pt]   [align=left] {\begin{minipage}[lt]{21.43pt}\setlength\topsep{0pt}
\begin{center}
\textbf{$\mathbb{R}_{c,i}$}
\end{center}

\end{minipage}};
% Text Node
\draw (356,7196) node [anchor=north west][inner sep=0.75pt]   [align=left] {\textit{\textbf{extrinsic}}};
% Text Node
\draw (360,7168) node [anchor=north west][inner sep=0.75pt]   [align=left] {\begin{minipage}[lt]{31.43pt}\setlength\topsep{0pt}
\begin{center}
\textbf{$\mathbb{R}_{w}$}
\end{center}

\end{minipage}};
% Text Node
\draw (316,7208.9) node [anchor=north west][inner sep=0.75pt]   [align=left] {\begin{minipage}[lt]{15.76pt}\setlength\topsep{0pt}
\begin{flushright}
\textbf{$\mathbb{R}_{s}$}
\end{flushright}

\end{minipage}};
% Text Node
\draw (426.36,7208.62) node [anchor=north west][inner sep=0.75pt]   [align=left] {\textbf{$\mathbb{R}_{c}$}};
% Text Node
\draw (247,7197.33) node [anchor=north west][inner sep=0.75pt]   [align=left] {\textbf{Sensor}};
% Text Node
\draw (471,7193.33) node [anchor=north west][inner sep=0.75pt]   [align=left] {\textbf{Camera}};

\end{tikzpicture}
 
    \caption{The pyramid-shaped illustration shows the structure of sensor and camera-related inertial systems. Information from different sensors of an intrinsic inertial system [$\mathbb{R}_{s,i}$,$\mathbb{R}_{c,i}$] can be transferred to extrinsic inertial systems [$\mathbb{R}_{s}$,$\mathbb{R}_{s}$] and a higher-level world coordinate system $\mathbb{R}_{w}$ via referenceable variables. }
    \label{fig:CameraSensorInertialSystem}
\end{figure}

Fig.~\ref{fig:CameraSensorInertialSystem} shows the schematic structure to transmit intrinsic sensor data to a higher-level inertial system.

\vspace*{-0.2cm}

\vspace*{-0.3cm}

The raw camera and IMU data refer to an intrinsic inertial system. Using a reference such as visual fixed points~\cite{5651059, 8793604}, GPS signals~\cite{10.1109/TITS.2012.2187641} or magnetic reference fields~\cite{Shi_2021, MK.22}, the data can be aligned and transferred to a higher-level world coordinate system $\mathbb{R}_{w}$. Generally, this refers to the Earth-Fixed Coordinate System (ECEF), expressed as $\mathbb{R}_{w}$ (shown in Fig.~\ref{fig:CameraSensorInertialSystem}). 

\newpage

To estimate the positions of robot systems it is necessary to transform the independent sensor types $\mathbb{R}_{s,i}$,$\mathbb{R}_{c,i}$ from intrinsic coordinate system to a higher-level world coordinate system $\mathbb{R}_{w}$, which serves as a unified, inertial system. This world coordinate system enables the correlation of different data. 

% Die Prinzip der relativistischen Bildverarbeitung beschränkt sich womöglich nicht nur auf Kameradaten. Es ist zu erproben inwiefern Event-basierte, Thermo oder Infrarotbilddaten anwendbar sind auf das Prinzip.

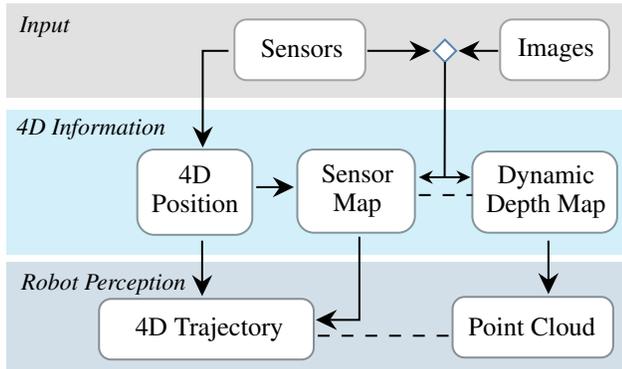
\begin{figure}[ht!]
    \centering

\tikzset{every picture/.style={line width=0.75pt}} %set default line width to 0.75pt        

\begin{tikzpicture}[x=0.75pt,y=0.75pt,yscale=-1,xscale=1]
%uncomment if require: \path (0,8452); %set diagram left start at 0, and has height of 8452

%Shape: Rectangle [id:dp23790643142718282] 
\draw  [draw opacity=0][fill={rgb, 255:red, 5; green, 76; blue, 126 }  ,fill opacity=0.18 ] (140.3,6546.43) -- (454.3,6546.43) -- (454.3,6602.23) -- (140.3,6602.23) -- cycle ;
%Shape: Rectangle [id:dp34408167784605315] 
\draw  [draw opacity=0][fill={rgb, 255:red, 155; green, 155; blue, 155 }  ,fill opacity=0.3 ] (140.3,6415.83) -- (454.3,6415.83) -- (454.3,6463.43) -- (140.3,6463.43) -- cycle ;
%Shape: Rectangle [id:dp34536569134936657] 
\draw  [draw opacity=0][fill={rgb, 255:red, 8; green, 168; blue, 232 }  ,fill opacity=0.18 ] (140.3,6469.43) -- (454.3,6469.43) -- (454.3,6542.43) -- (140.3,6542.43) -- cycle ;
%Rounded Rect [id:dp7018522723999785] 
\draw  [color={rgb, 255:red, 0; green, 0; blue, 0 }  ,draw opacity=0.28 ][fill={rgb, 255:red, 255; green, 255; blue, 255 }  ,fill opacity=1 ] (255.3,6429.95) .. controls (255.3,6426.57) and (258.04,6423.83) .. (261.42,6423.83) -- (315.18,6423.83) .. controls (318.56,6423.83) and (321.3,6426.57) .. (321.3,6429.95) -- (321.3,6448.31) .. controls (321.3,6451.69) and (318.56,6454.43) .. (315.18,6454.43) -- (261.42,6454.43) .. controls (258.04,6454.43) and (255.3,6451.69) .. (255.3,6448.31) -- cycle ;

%Rounded Rect [id:dp14625937633325736] 
\draw  [color={rgb, 255:red, 0; green, 0; blue, 0 }  ,draw opacity=0.28 ][fill={rgb, 255:red, 255; green, 255; blue, 255 }  ,fill opacity=1 ] (389.3,6429.95) .. controls (389.3,6426.57) and (392.04,6423.83) .. (395.42,6423.83) -- (438.18,6423.83) .. controls (441.56,6423.83) and (444.3,6426.57) .. (444.3,6429.95) -- (444.3,6448.31) .. controls (444.3,6451.69) and (441.56,6454.43) .. (438.18,6454.43) -- (395.42,6454.43) .. controls (392.04,6454.43) and (389.3,6451.69) .. (389.3,6448.31) -- cycle ;
%Rounded Rect [id:dp3282627176546793] 
\draw  [color={rgb, 255:red, 0; green, 0; blue, 0 }  ,draw opacity=0.36 ][fill={rgb, 255:red, 255; green, 255; blue, 255 }  ,fill opacity=1 ] (365.3,6569.95) .. controls (365.3,6566.57) and (368.04,6563.83) .. (371.42,6563.83) -- (440.18,6563.83) .. controls (443.56,6563.83) and (446.3,6566.57) .. (446.3,6569.95) -- (446.3,6588.31) .. controls (446.3,6591.69) and (443.56,6594.43) .. (440.18,6594.43) -- (371.42,6594.43) .. controls (368.04,6594.43) and (365.3,6591.69) .. (365.3,6588.31) -- cycle ;
%Rounded Rect [id:dp3974166315564165] 
\draw  [color={rgb, 255:red, 0; green, 0; blue, 0 }  ,draw opacity=0.36 ][fill={rgb, 255:red, 255; green, 255; blue, 255 }  ,fill opacity=1 ] (187,6570.95) .. controls (187,6567.57) and (189.74,6564.83) .. (193.12,6564.83) -- (289.18,6564.83) .. controls (292.56,6564.83) and (295.3,6567.57) .. (295.3,6570.95) -- (295.3,6589.31) .. controls (295.3,6592.69) and (292.56,6595.43) .. (289.18,6595.43) -- (193.12,6595.43) .. controls (189.74,6595.43) and (187,6592.69) .. (187,6589.31) -- cycle ;

%Rounded Rect [id:dp12916972658057424] 
\draw  [color={rgb, 255:red, 0; green, 0; blue, 0 }  ,draw opacity=0.38 ][fill={rgb, 255:red, 255; green, 255; blue, 255 }  ,fill opacity=1 ] (206.3,6498.04) .. controls (206.3,6493.28) and (210.16,6489.43) .. (214.91,6489.43) -- (255.69,6489.43) .. controls (260.44,6489.43) and (264.3,6493.28) .. (264.3,6498.04) -- (264.3,6523.88) .. controls (264.3,6528.63) and (260.44,6532.49) .. (255.69,6532.49) -- (214.91,6532.49) .. controls (210.16,6532.49) and (206.3,6528.63) .. (206.3,6523.88) -- cycle ;
%Straight Lines [id:da4516403037001171] 
\draw    (252.3,6439.63) -- (239.3,6439.63) -- (239.3,6485.43) ;
\draw [shift={(239.3,6488.43)}, rotate = 270] [fill={rgb, 255:red, 0; green, 0; blue, 0 }  ][line width=0.08]  [draw opacity=0] (10.72,-5.15) -- (0,0) -- (10.72,5.15) -- (7.12,0) -- cycle    ;
%Straight Lines [id:da9761123835912657] 
\draw    (239,6535.43) -- (239,6560.43) ;
\draw [shift={(239,6563.43)}, rotate = 270] [fill={rgb, 255:red, 0; green, 0; blue, 0 }  ][line width=0.08]  [draw opacity=0] (10.72,-5.15) -- (0,0) -- (10.72,5.15) -- (7.12,0) -- cycle    ;
%Straight Lines [id:da16612514912006593] 
\draw    (413.65,6535.43) -- (413.65,6559.43) ;
\draw [shift={(413.65,6562.43)}, rotate = 270] [fill={rgb, 255:red, 0; green, 0; blue, 0 }  ][line width=0.08]  [draw opacity=0] (10.72,-5.15) -- (0,0) -- (10.72,5.15) -- (7.12,0) -- cycle    ;
%Straight Lines [id:da34386797771752864] 
\draw    (361.3,6503.43) -- (351.3,6503.43) ;
\draw [shift={(348.3,6503.43)}, rotate = 360] [fill={rgb, 255:red, 0; green, 0; blue, 0 }  ][line width=0.08]  [draw opacity=0] (7.14,-3.43) -- (0,0) -- (7.14,3.43) -- (4.74,0) -- cycle    ;
%Straight Lines [id:da2670463633356399] 
\draw    (266.3,6508.43) -- (282.3,6508.43) ;
\draw [shift={(285.3,6508.43)}, rotate = 180] [fill={rgb, 255:red, 0; green, 0; blue, 0 }  ][line width=0.08]  [draw opacity=0] (10.72,-5.15) -- (0,0) -- (10.72,5.15) -- (7.12,0) -- cycle    ;
%Straight Lines [id:da5680574060968513] 
\draw  [dash pattern={on 4.5pt off 4.5pt}]  (348.3,6512.43) -- (375.3,6512.43) ;
%Straight Lines [id:da8251005281706194] 
\draw  [dash pattern={on 4.5pt off 4.5pt}]  (297.3,6583.43) -- (367.3,6583.43) ;
%Shape: Diamond [id:dp2708610751995508] 
\draw  [color={rgb, 255:red, 10; green, 78; blue, 146 }  ,draw opacity=0.64 ][fill={rgb, 255:red, 255; green, 255; blue, 255 }  ,fill opacity=1 ] (361.45,6433.98) -- (367.1,6439.63) -- (361.45,6445.28) -- (355.8,6439.63) -- cycle ;
%Rounded Rect [id:dp9099903396208591] 
\draw  [color={rgb, 255:red, 0; green, 0; blue, 0 }  ,draw opacity=0.38 ][fill={rgb, 255:red, 255; green, 255; blue, 255 }  ,fill opacity=1 ] (287,6497.83) .. controls (287,6493.19) and (290.76,6489.43) .. (295.4,6489.43) -- (337.9,6489.43) .. controls (342.54,6489.43) and (346.3,6493.19) .. (346.3,6497.83) -- (346.3,6523.03) .. controls (346.3,6527.66) and (342.54,6531.43) .. (337.9,6531.43) -- (295.4,6531.43) .. controls (290.76,6531.43) and (287,6527.66) .. (287,6523.03) -- cycle ;
%Rounded Rect [id:dp37267756956056264] 
\draw  [color={rgb, 255:red, 0; green, 0; blue, 0 }  ,draw opacity=0.38 ][fill={rgb, 255:red, 255; green, 255; blue, 255 }  ,fill opacity=1 ] (375.3,6498.63) .. controls (375.3,6494.1) and (378.97,6490.43) .. (383.5,6490.43) -- (440.1,6490.43) .. controls (444.63,6490.43) and (448.3,6494.1) .. (448.3,6498.63) -- (448.3,6523.23) .. controls (448.3,6527.75) and (444.63,6531.43) .. (440.1,6531.43) -- (383.5,6531.43) .. controls (378.97,6531.43) and (375.3,6527.75) .. (375.3,6523.23) -- cycle ;
%Straight Lines [id:da9426569473219908] 
\draw    (322.3,6439.63) -- (351.2,6439.63) ;
\draw [shift={(354.2,6439.63)}, rotate = 180] [fill={rgb, 255:red, 0; green, 0; blue, 0 }  ][line width=0.08]  [draw opacity=0] (10.72,-5.15) -- (0,0) -- (10.72,5.15) -- (7.12,0) -- cycle    ;
%Straight Lines [id:da8648084941851677] 
\draw    (388.3,6439.63) -- (371.3,6439.63) ;
\draw [shift={(368.3,6439.63)}, rotate = 360] [fill={rgb, 255:red, 0; green, 0; blue, 0 }  ][line width=0.08]  [draw opacity=0] (10.72,-5.15) -- (0,0) -- (10.72,5.15) -- (7.12,0) -- cycle    ;
%Straight Lines [id:da05119530151759233] 
\draw    (361.45,6445.28) -- (361.3,6503.43) -- (371.3,6503.43) ;
\draw [shift={(374.3,6503.43)}, rotate = 180] [fill={rgb, 255:red, 0; green, 0; blue, 0 }  ][line width=0.08]  [draw opacity=0] (7.14,-3.43) -- (0,0) -- (7.14,3.43) -- (4.74,0) -- cycle    ;

%Straight Lines [id:da08514096084786549] 
\draw    (318.3,6533.43) -- (318.3,6575.43) -- (299.3,6575.43) ;
\draw [shift={(296.3,6575.43)}, rotate = 360] [fill={rgb, 255:red, 0; green, 0; blue, 0 }  ][line width=0.08]  [draw opacity=0] (10.72,-5.15) -- (0,0) -- (10.72,5.15) -- (7.12,0) -- cycle    ;

% Text Node
\draw (144,6550.43) node [anchor=north west][inner sep=0.75pt]   [align=left] {\begin{minipage}[lt]{63.98pt}\setlength\topsep{0pt}
\begin{flushright}
\textit{{\footnotesize Robot Perception}}
\end{flushright}

\end{minipage}};
% Text Node
\draw (140.3,6420.83) node [anchor=north west][inner sep=0.75pt]   [align=left] {\begin{minipage}[lt]{23pt}\setlength\topsep{0pt}
\begin{flushright}
\textit{{\footnotesize Input}}
\end{flushright}

\end{minipage}};
% Text Node
\draw (208,6495.82) node [anchor=north west][inner sep=0.75pt]   [align=left] {\begin{minipage}[lt]{39.02pt}\setlength\topsep{0pt}
\begin{center}
4D \\Position
\end{center}

\end{minipage}};
% Text Node
\draw (290.55,6495.34) node [anchor=north west][inner sep=0.75pt]   [align=left] {\begin{minipage}[lt]{37.88pt}\setlength\topsep{0pt}
\begin{center}
Sensor \\Map
\end{center}

\end{minipage}};
% Text Node
\draw (374,6496) node [anchor=north west][inner sep=0.75pt]   [align=left] {\begin{minipage}[lt]{55.45pt}\setlength\topsep{0pt}
\begin{center}
Dynamic\\ Depth Map
\end{center}

\end{minipage}};
% Text Node
\draw (144,6472.43) node [anchor=north west][inner sep=0.75pt]   [align=left] {\textit{{\footnotesize 4D Information}}};
% Text Node
\draw (387,6432) node [anchor=north west][inner sep=0.75pt]   [align=left] {\begin{minipage}[lt]{36.18pt}\setlength\topsep{0pt}
\begin{flushright}
Images
\end{flushright}

\end{minipage}};
% Text Node
\draw (362,6573) node [anchor=north west][inner sep=0.75pt]   [align=left] {\begin{minipage}[lt]{55.45pt}\setlength\topsep{0pt}
\begin{flushright}
Point Cloud
\end{flushright}

\end{minipage}};
% Text Node
\draw (194,6573) node [anchor=north west][inner sep=0.75pt]   [align=left] {\begin{minipage}[lt]{63.38pt}\setlength\topsep{0pt}
\begin{flushright}
4D Trajectory
\end{flushright}

\end{minipage}};
% Text Node
\draw (255,6432) node [anchor=north west][inner sep=0.75pt]   [align=left] {\begin{minipage}[lt]{40.15pt}\setlength\topsep{0pt}
\begin{flushright}
Sensors
\end{flushright}

\end{minipage}};

\end{tikzpicture}

    \caption{\textbf{Design of 4D-based Robot Perception} Robot perception contains the 4D trajectory of own position and the relation of point clouds for the perception of surrounding distances. The 4D information can be extracted from various sensors as positional and visual information.    }
    \label{fig:RobotPerception}
\end{figure}

By extracting 4D information in terms of 4D positioning, sensor maps, and dynamic-dependent depth maps, robot perception can be extended to 4D-based trajection and ambient point clouds to navigate robot systems, shown in Fig.~\ref{fig:RobotPerception}. 

% \vspace*{-0.8cm}

Based on relativistic sensor processing, environment-related sensor and image data can be mapped over time as 4D information. This enables the visualization of sensor maps and depth maps through the fusion of perspective changes in images and the corresponding sensor information, which are associated with the dynamic change of a measurable environment. Robot's perception can include a 4D trajectory as a part of its position, current and predictive path planning as well as environment-related sensor information such as forces, velocities, or accelerations as vector lengths.

\section{Discussion}
\begin{figure}[ht!]
    \centering
    \includegraphics[width=0.95\linewidth]{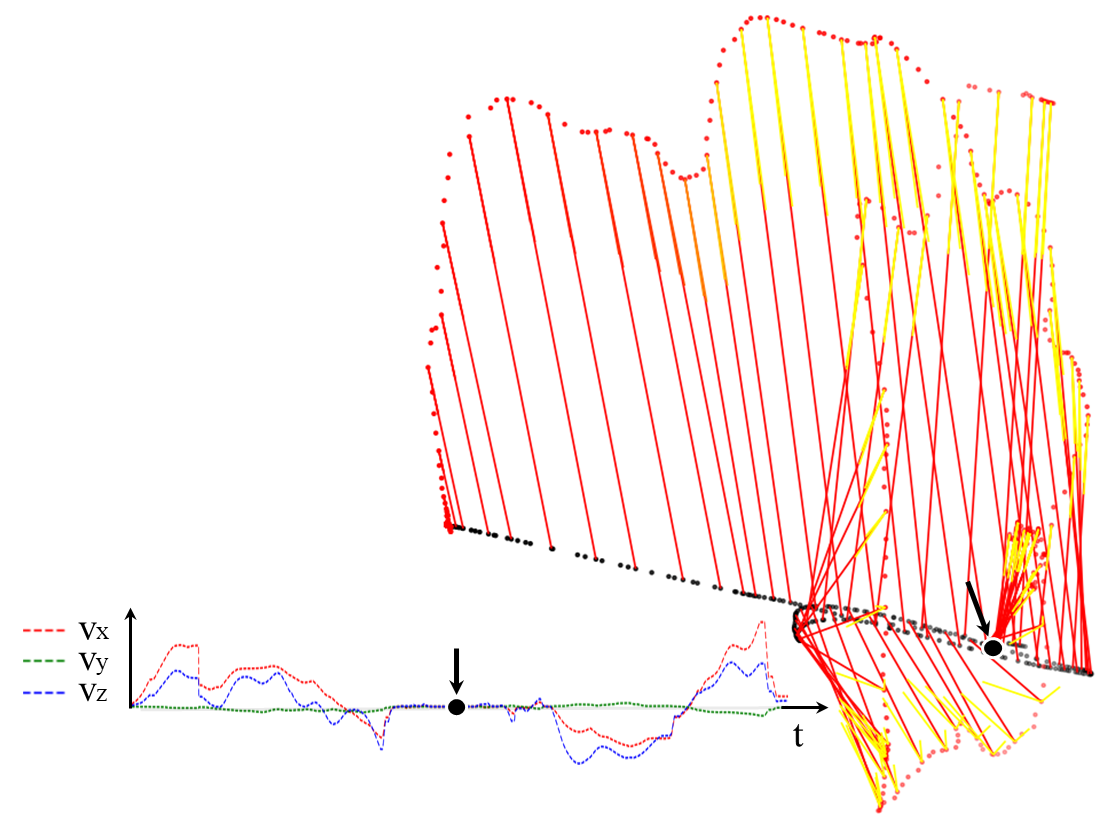}
    \vspace*{-0.3cm}
    \caption{The illustrated motion of 4D-trajectory includes the sensor orientation and the velocity level. The yellow arrows refer to the relatively determined maximum speed. The arrow length is a further indicator of temporal-related velocity. If no arrow is present, the sensor is in static state ($v=0$). Applying the Schlingel diagram, the 4D position can be related to the velocity-dependent vector length.    }
    \label{fig:SchlingelDarstellungV2}
\end{figure}

Our model estimates positions and distances over space and time by extracting various types of environmental information. Considering the information density, the Schlingel representation can contain further information like velocities or forces, shown in Fig.~\ref{fig:SchlingelDarstellungV2}.  

In the Schlingel diagram, the information are related to the immediate surroundings over time. In this way, sensor-dependent variables like forces, velocity, or accelerations can be visualized in vectorial orientation and length. This offers the possibility of correcting position deviations with environmental sensor data that are indirectly related to the used sensor information of our system. Further sensor variables can be placed in a similar relationship using the Schlingel diagram.

\newpage
%\vspace*{-0.3cm}

\begin{figure}[ht!]
    \centering
    \includegraphics[width=1\linewidth]{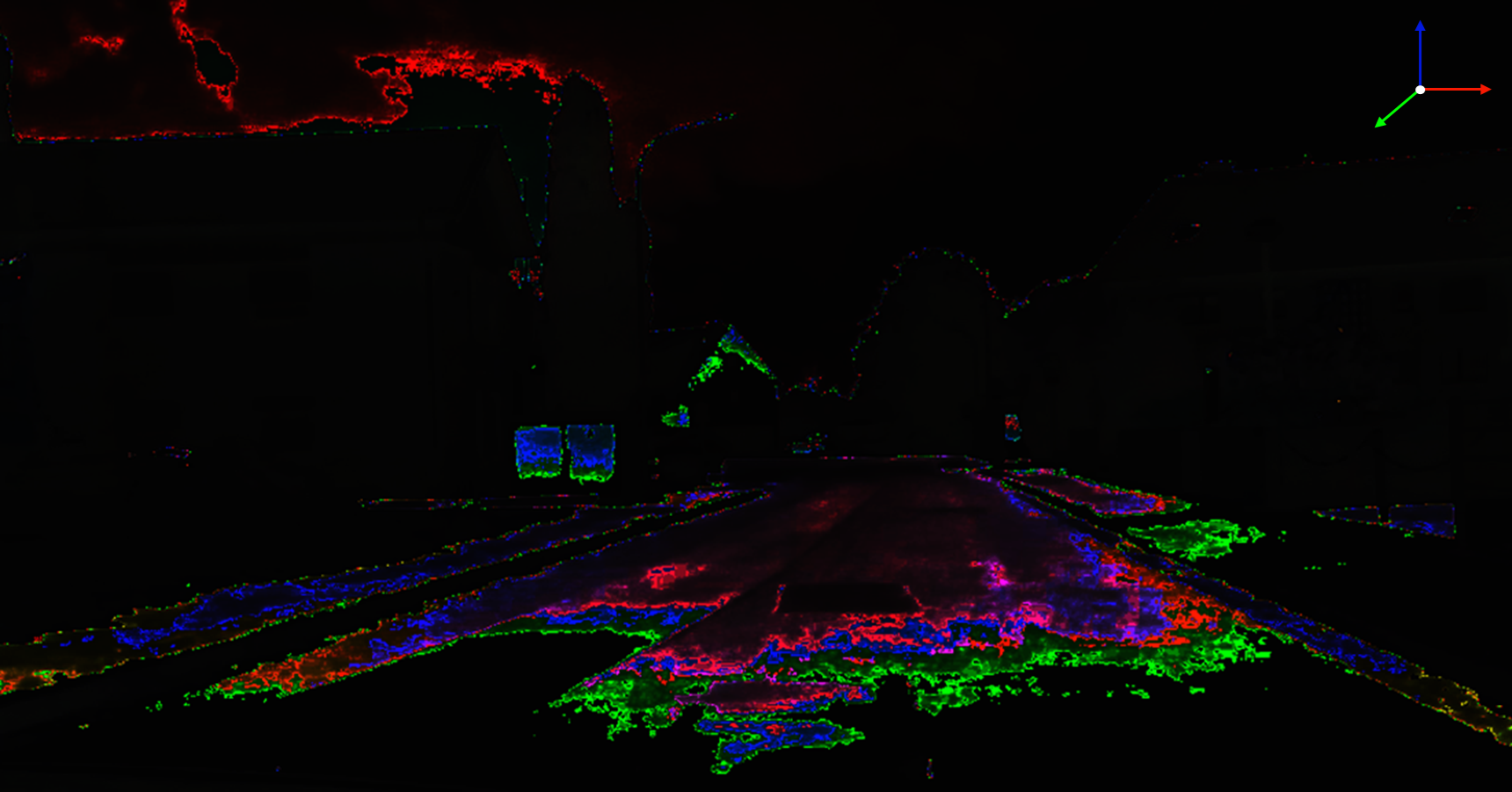}
    \caption{The illustration shows a sensor map which is generated by image and sensor data. The different colors indicate the respective positional change in space. As part of relativistic image processing, the sensory map is suitable for visual estimation of distances and can be extended to robot navigation.}
    \label{fig:SensorMap}
\end{figure}

The 4D model enables the extended visualization of additional information as it refers to the respective and environment-specific position data of various dynamic variables. Position changes can be illustrated by our 4D model as a kind of sensor map, extracted by images (Fig.~\ref{fig:SensorMap}).

\vspace*{-0.0cm}

The illustrated sensor map reveals the influence of acceleration sensors in the form of translational change in velocity. The different colors indicate the respective positional change of motion in space. By combining color and diagram-specific position data using relativistic image processing, navigation of robot systems can be extended to further sensor information. This can improve the accuracy and robustness of these systems without using direct AI approaches. 

\newpage

 Combining 4D trajectories and sensor maps favors the navigation of robotic systems. For example, local-temporal sensor information and visual sensor maps can be used as a reference to initialize the positions for navigation. In addition, existing approaches such as geomagnetic inertial navigation~\cite{MK.22}, in which the earth's magnetic field is used as a reference, can be combined with the 4D navigation described above.  

\section{Conclusion}

This paper presents a novel concept of 4D-based navigation of robot systems. Our motivation is based on different challenges in a dynamic environment where weather, light, or contrast conditions can vary and motion is dependent on time. To estimate the 4D position and related distances in spatio-temporal view, we refer to the Schlingel diagram as a 4D model with 4 translations and 6 rotations. The method of 4D-Navigation relates to the principles of relativistic image processing. 

The 4D-based navigation of robotic systems proves to be a promising approach. The processing of acquired sensor information in a unit tensor field using the Schlingel diagram enables the integration of spatial and temporal information. Furthermore, this model refers to the dependency of several and also different sensor types in the same environment. The failure or inaccuracy of individual sensors can thus be significantly reduced. 

We evaluated our results by acquiring acceleration, angular velocity, and magnetic field of an environment in a moving scene. The images have been extracted from a stereoscopic camera. The test set of data is limited to a few data sets and supports the basic functionality. In future work, more and more diverse data sets will be collected. This also includes the expansion of sensor scope in terms of Time-of-Flight (ToF), LiDAR, event-based cameras, 360-degree cameras, as well as pressure, temperature, and haptic sensors. 

Our concept of 4D-based navigation describes an equation-specific and sensor-based model of navigation. In the future, we intend to transfer our model to a diversity of navigation and AI approaches such as inertial navigation, magnetic inertial navigation, VINS, and learning-based navigation for robot control. This involves the verification of how reliable the system is due to the 4D model and how interchangeable the sensors are.

\bibliography{main}

\begin{thebibliography}{18}
\providecommand{\natexlab}[1]{#1}

\bibitem[{Boblest, Müller, and Wunner(2022)}]{BMW.22}
Boblest, S.; Müller, T.; and Wunner, G. 2022.
\newblock Spezielle und allgemeine Relativitätstheorie.
\newblock Springer.

\bibitem[{Condurache(2022)}]{D.22}
Condurache, D. 2022.
\newblock Higher-Order Relative Kinematics of Rigid Body and Multibody Systems. A Novel Approach with Real and Dual Lie Algebras.
\newblock Elsevier - Mechanism and Machine Theory.

\bibitem[{Einstein(1916)}]{E.16}
Einstein, A. 1916.
\newblock Die Grundlage der allgemeinen Relativitätstheorie.
\newblock Wiley.

\bibitem[{Erickson and LaValle(2013)}]{Erickson13}
Erickson, L.; and LaValle, S. 2013.
\newblock A Simple, but NP-Hard, Motion Planning Problem.
\newblock Proceedings of the AAAI Conference on Artificial Intelligence.

\bibitem[{Garmin(Visited on August 2024)}]{gar22}
Garmin. Visited on August 2024.
\newblock Garmin Tracking System - Oregon 700.

\bibitem[{Huang(2019)}]{8793604}
Huang, G. 2019.
\newblock Visual-Inertial Navigation: A Concise Review.
\newblock In \emph{2019 International Conference on Robotics and Automation (ICRA)}, 9572--9582.

\bibitem[{Husqvarna(Visited on August 2024)}]{husqvarna24}
Husqvarna. Visited on August 2024.
\newblock Husqvarna automated and mobile machine.

\bibitem[{Kadambi, Bhandari, and Raskar(2014)}]{Kadambi20143DDC}
Kadambi, A.; Bhandari, A.; and Raskar, R. 2014.
\newblock 3D Depth Cameras in Vision: Benefits and Limitations of the Hardware.
\newblock Scopus.

\bibitem[{Kok, Hol, and Schön(2018)}]{kok18}
Kok, M.; Hol, J.~D.; and Schön, T.~B. 2018.
\newblock Using Inertial Sensors for Position and Orientation Estimation.
\newblock Foundations and Trends in Signal Processing.

\bibitem[{Loganathan and Ahmad(2023)}]{LOGANATHAN2023101343}
Loganathan, A.; and Ahmad, N.~S. 2023.
\newblock A systematic review on recent advances in autonomous mobile robot navigation.
\newblock volume~40, 101343.

\bibitem[{Martial et~al.(2016)Martial, Martial, Besnerais, and Vissiere}]{car16}
Martial, D.~C.; Martial, S.; Besnerais, G.~L.; and Vissiere, D. 2016.
\newblock Infrastructureless Indoor Navigation With an Hybrid Magneto-inertial and Depth Sensor System.
\newblock International Conference on Indoor Positioning and Indoor Navigation (IPIN).

\bibitem[{Müller and Kranzlmüller(2022)}]{MK.22}
Müller, S.; and Kranzlmüller, D. 2022.
\newblock Dynamic Sensor Matching based on Geomagnetic Inertial Navigation.
\newblock International Conference in Central Europe on Computer Graphics, Visualization and Computer Vision.

\bibitem[{Rahaman(2022)}]{Farook2022}
Rahaman, F. 2022.
\newblock The Special Theory of Relativity: A Mathematical Approach.
\newblock Springer Singapore.

\bibitem[{Ram{\'i}k et~al.(2013)Ram{\'i}k, Sabourin, Moreno, and Madani}]{RSMM.13}
Ram{\'i}k, D.~M.; Sabourin, C.; Moreno, R.; and Madani, K. 2013.
\newblock A machine learning based intelligent vision system for autonomous object detection and recognition.
\newblock volume~40, 358 -- 375.

\bibitem[{Shi et~al.(2021)Shi, Gao, Gao, Ding, and Zhang}]{Shi_2021}
Shi, S.; Gao, T.; Gao, D.; Ding, Z.; and Zhang, Z. 2021.
\newblock Inertial navigation aid indoor navigation based on the establishment of accurate magnetic reference map.
\newblock volume 1802, 042022. IOP Publishing.

\bibitem[{Siciliano and Khatib(2016)}]{SK.16}
Siciliano, B.; and Khatib, O. 2016.
\newblock Springer Handbook of Robotics.
\newblock Springer.

\bibitem[{Tardif et~al.(2010)Tardif, George, Laverne, Kelly, and Stentz}]{5651059}
Tardif, J.-P.; George, M.; Laverne, M.; Kelly, A.; and Stentz, A. 2010.
\newblock A new approach to vision-aided inertial navigation.
\newblock In \emph{2010 IEEE/RSJ International Conference on Intelligent Robots and Systems}, 4161--4168.

\bibitem[{Vu et~al.(2012)Vu, Ramanandan, Chen, Farrell, and Barth}]{10.1109/TITS.2012.2187641}
Vu, A.; Ramanandan, A.; Chen, A.; Farrell, J.~A.; and Barth, M. 2012.
\newblock Real-Time Computer Vision/DGPS-Aided Inertial Navigation System for Lane-Level Vehicle Navigation.
\newblock volume~13, 899–913. IEEE Press.

\end{thebibliography}

\end{document}